%% file: main.tex
\documentclass[11pt]{article}

\usepackage[final]{acl}

\usepackage{times}
\usepackage{latexsym}
\usepackage[T1]{fontenc}
\usepackage[utf8]{inputenc}
\usepackage{microtype}
\usepackage{inconsolata}
\usepackage{graphicx}
\usepackage{multirow}
\usepackage{enumitem}
\usepackage{xspace}
\usepackage[table]{xcolor}
\usepackage[most]{tcolorbox}
\usepackage{listings}
\usepackage{booktabs}
\usepackage{comment}
\usepackage{amsmath}
\usepackage{amssymb}

\lstdefinestyle{jsonseed}{
    basicstyle=\ttfamily\footnotesize,
    breaklines=true,                 
    breakatwhitespace=false,        
    breakindent=0pt,
    postbreak=\mbox{\textcolor{gray}{$\hookrightarrow$}\space},
    columns=fullflexible,
    keepspaces=true,
    showstringspaces=false,
    upquote=true,
}

\title{Exploring Autonomous Agentic Data Engineering \\ for Model Specialization}

\author{
  \textbf{Yujie Luo}$^{\spadesuit\heartsuit}$\thanks{Equal Contribution.},
  \textbf{Xiangyuan Ru}$^{\spadesuit}$\footnotemark[1],
  \textbf{Jingsheng Zheng}$^{\spadesuit}$, 
  \textbf{Jingjing Wang}$^{\spadesuit}$, 
  \textbf{Yuqi Zhu}$^{\spadesuit}$, \\
  \textbf{Jintian Zhang}$^{\spadesuit}$, 
  \textbf{Runnan Fang}$^{\spadesuit}$, 
  \textbf{Kewei Xu}$^{\spadesuit}$, 
  \textbf{Ye Liu}$^{\heartsuit}$, 
  \textbf{Zheng Wei}$^{\heartsuit}$\thanks{Corresponding Authors.}, \\
  \textbf{Jiang Bian}$^{\heartsuit}$, 
  \textbf{Zang Li}$^{\heartsuit}$, 
  \textbf{Shumin Deng}$^{\spadesuit,\clubsuit}$\footnotemark[\value{footnote}] 
  \\
  $^\spadesuit$Zhejiang University 
  \quad
  $^\heartsuit$Platform and Content Group, Tencent 
  \\
  \textsuperscript{$\clubsuit$}State Key Lab. of Novel Software Technology,  Nanjing University, P.R. China
  \\
  \texttt{\{luo.yj,231sm\}@zju.edu.cn}
}

\begin{document}
\maketitle

\begin{abstract}
Large Language Models (LLMs) have demonstrated strong performance on general tasks, while often struggling to adapt to specialized domains without high-quality domain-specific data. 
Existing LLM-based data curation methods primarily rely on human-designed workflows, leaving it unexamined whether LLMs can autonomously execute an end-to-end data engineering pipeline for model specialization.
We formalize \textbf{Autonomous Agentic Data Engineering}, a novel task designed to evaluate LLMs as autonomous data engineers that drive model specialization through end-to-end data curation.
We frame data as an optimizable component and study agents that plan, generate, and iteratively optimize training data across multiple domains, guided by post-training performance improvement.
Experiments show that autonomous LLM data engineers yield substantial gains, as GPT-5.2 constructs a training curriculum that improves a student model by \textbf{57.29\%}, entirely through iterative, agent-driven data adaptation.
By illuminating both potential and bottlenecks, our study establishes autonomous data engineering as a measurable capability and charts a path toward agent-driven model specialization\footnote{Code at \url{https://github.com/zjunlp/DataAgent}.}.
\end{abstract}

\input{section/introduction}

\input{section/method}

\input{section/experiments}

\input{section/analysis}

\input{section/related_work}

\input{section/conclusion}

\section*{Limitations}

We acknowledge several limitations in our work. 
First, although focusing on QA tasks allows us to efficiently obtain reliable environmental feedback for closed-loop optimization, this design restricts our evaluation on open-ended generation tasks where automated evaluation is difficult to achieve. 
Second, despite implementing strict budget caps, the Iterative Agent still demands considerable computational resources for model inference and fine-tuning.
Finally, while we average the results across multiple runs to mitigate fluctuations, coupling complex end-to-end data engineering tasks still introduces unavoidable run-to-run variance. We leave broader task coverage and more cost-efficient strategies to future work.

\section*{Acknowledgements}
We would like to express our sincere gratitude to the anonymous reviewers for their thoughtful and constructive feedback. 
This work was supported by the New Generation Artificial Intelligence-National Science and Technology Major Project (2025ZD0122802), National Natural Science Foundation of China (No.62676362, No.NSFCU23B2055, No.NSFCU19B2027), the Fundamental Research Funds for the Central Universities (226-2023-00138), CCF-Tencent Open Research Fund, Tencent PCG Venus, the Yongjiang Talent Introduction Programme (2021A-156-G), the State Key Lab. of Novel Software Technology,  Nanjing University, P.R. China (KFKT2026B02), and the Information Technology Center and State Key Lab of CAD\&CG, Zhejiang University. 
This work was also supported by Xinmiao Project of Zhejiang Provincial Applied Basic Research Program (2026XMZD064).

\bibliography{reference}

\appendix

\input{appendix/dataset}

\input{appendix/exp_details}

\input{appendix/exp_settings}

\input{appendix/platform}

\input{appendix/student_teacher}

\input{appendix/leakage_analysis}

\input{appendix/generalize}

\input{appendix/improve_analysis}

\input{appendix/failure_analysis}

\input{appendix/seed_example}

\input{appendix/prompt}

\end{document}

%% file: section/introduction.tex
\section{Introduction}   

Large Language Models (LLMs) have acquired emergent capabilities through training on massive amounts of data 
\cite{DBLP:journals/corr/abs-2506-04178, DBLP:conf/infocom/ZhouXWLWCZL25} in recent years.
Despite strong performance on general tasks, even the most advanced LLMs often struggle to adapt when their training data do not adequately reflect specialized downstream tasks \cite{DBLP:conf/acl/LiTZSX24,DBLP:conf/acl/MishraKBH22}. 

Adapting a general-purpose model to a target specialized domain typically necessitates post-training on domain-specific instruction data, as exemplified by curated corpora \cite{DBLP:conf/nips/ZhangHZDYWYD024, DBLP:journals/corr/abs-2306-06031}. 
Given the complexity of data processing and the scarcity of high-quality domain data, researchers have increasingly turned to LLM-based methods \cite{DBLP:conf/acl/Qiao0FLZJLC24,liang2025dataflow}, utilizing LLMs as data generators within human-designed workflows.
As adapting these handcrafted recipes to new domains requires extensive configuration, modern LLM agents offer a more promising alternative through their remarkable advances in complex reasoning \cite{DBLP:journals/corr/abs-2501-12948}, code generation \cite{DBLP:conf/icml/Ni0RSYWL23, DBLP:conf/iclr/HongZCZCWZWYLZR24}, and tool use \cite{DBLP:conf/iclr/QinLYZYLLCTQZHT24}.
These advances further raise a natural question: \textit{\textbf{Can LLM agents autonomously perform end-to-end data engineering for model specialization?}}

\begin{figure}[t]
  \centering
  \includegraphics[width=0.9\linewidth]{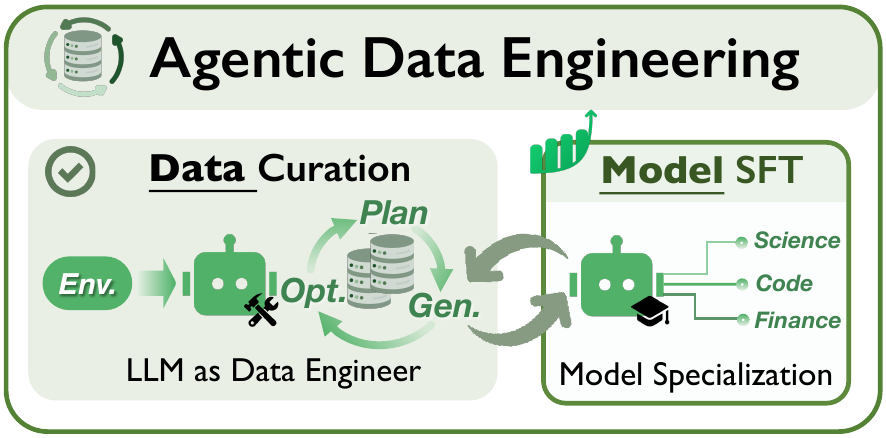}
  \caption{Paradigm of \textbf{Agentic Data Engineering}. LLM data engineer independently executes the entire data curation loop to drive model specialization, iteratively optimizing data guided by post-training student model performance feedback. }
  \vspace{-4mm}
  \label{fig:brief}
\end{figure}

To investigate this question, we formalize the task of \textbf{Autonomous Agentic Data Engineering} (Figure \ref{fig:brief}), where LLMs are tasked with completing the entire training data curation pipeline independently, including strategy plan, domain specification, prompt design, data synthesis, data validation, and iterative data optimization. 
By holding both the teacher model for data synthesis and the student model for data training fixed, we isolate the end-to-end data engineering capability of LLMs, which is ultimately evaluated by the post-training performance improvement of the student model.

We conduct a comprehensive analysis of the performance of mainstream LLMs across three specialized domains: Science, Code, and Finance. 
LLM capabilities are evaluated under a single-turn completion agent setting (One-Shot) and a closed-loop, self-optimizing agent setting (Iterative Agent), both from scratch and with initial seed data. 
Experiments show that modern LLM agents possess substantial data engineering capabilities, enabling them to infer missing supervision signals and synthesize task-aligned instances even from scratch. 
Notably, GPT-5.2 achieves an average relative \emph{performance gain of} \textbf{57.29\%} through iterative optimization, surpassing human-crafted data synthesis pipelines. 
Despite these encouraging findings, we also identify significant failure modes, suggesting that LLMs still lack robust post-generation mechanisms for reliable quality assurance.

Overall, we summarize our contributions as:
\begin{itemize}[leftmargin=*]
    \item We formalize the task of \textbf{Agentic Data Engineering}, an autonomous paradigm in which LLMs independently manage the entire training data curation lifecycle. This provides a controlled setting for studying end-to-end data engineering as a measurable capability of LLM agents.
    
    \item We develop an end-to-end execution \& evaluation environment that covers the full data curation pipeline for model specialization, enabling isolated and budget-controlled agent execution, along with external feedback and a performance-based evaluation protocol.
    
    \item We instantiate two representative settings: \textit{One-Shot} and \textit{Iterative Agent}, and evaluate mainstream LLMs across diverse domains. We further provide analysis of iterative optimization, data quality, and failure modes towards specialization.
\end{itemize}

%% file: section/method.tex
\section{Agentic Data Engineering}

\begin{figure*}[t]
  \centering
  \includegraphics[width=0.98\linewidth]{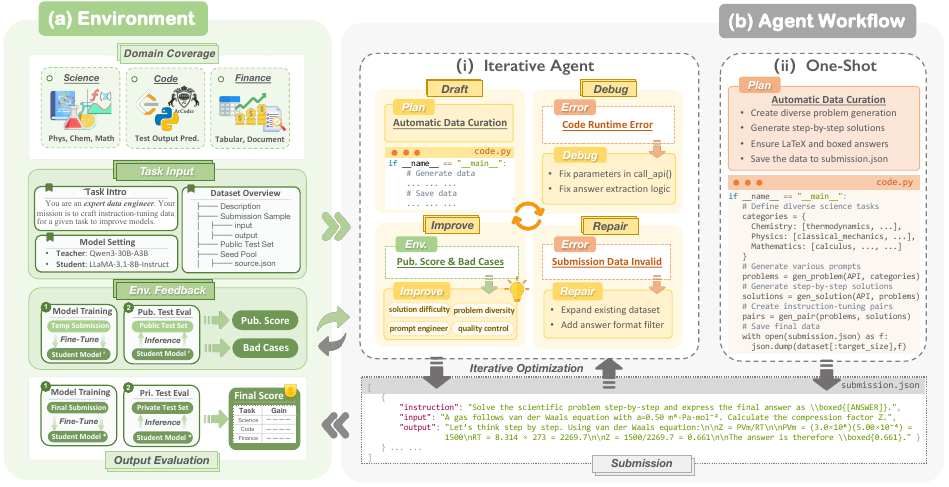}
  \vspace{-3mm}
  \caption{\textbf{Overall framework of our study}. \textbf{(a) Environment}: the overview of the covered domains, the agent input containing task settings and procedural feedback, and the final evaluation method. \textbf{(b) Agent Workflow}: the example workflow in which agents develop strategies to curate data and output a \texttt{submission.json} towards specialization. In (ii) One-Shot setting, the submission is produced in a single pass, whereas in (i) Iterative Agent setting, the agent iteratively improves its data curation strategy with feedback and reports the best submission.}
  \label{fig:main}
  \vspace{-3mm}
\end{figure*}

\label{sec:autodatabench}
\subsection{Problem Formulation}
\label{subsec:task_formulation}

We formalize \textbf{Agentic Data Engineering} (Figure \ref{fig:brief}) as an end-to-end closed-loop paradigm in which an LLM agent $\mathcal{A}$ autonomously curates training data to specialize a \emph{fixed} student model $\mathcal{M}_S$ with a \emph{fixed} teacher model $\mathcal{M}_T$ for data synthesis. 

For a target task $\mathcal{T}$, the agent designs a data-curation program $\mathcal{P}_{\mathcal{A}}$ that calls $\mathcal{M}_T$ to synthesize a candidate dataset
\begin{equation}
\widehat{\mathcal{D}} \;=\; \mathcal{P}_{\mathcal{A}}(\mathcal{T};\,\mathcal{M}_T).
\end{equation}
The student model is then specialized on $\widehat{\mathcal{D}}$ via supervised fine-tuning, denoted $\mathrm{Spec}(\cdot)$, and scored by a deterministic rule-based evaluator $\mathcal{E}$, producing the environmental feedback signal
\begin{equation}
f \;=\; \mathcal{E}\bigl(\mathrm{Spec}(\mathcal{M}_S,\,\widehat{\mathcal{D}})\bigr).
\end{equation}

Given the synthesis data $\widehat{\mathcal{D}}$ and the feedback signal $f$, the entire agentic data engineering process can be cast as a closed-loop objective in which agent $\mathcal{A}$ searches over curation strategies to maximize the student's post-training performance:
\begin{equation}
\mathcal{P}_{\mathcal{A}}^{\star} \;=\; \arg\max_{\mathcal{P}_{\mathcal{A}}}\;
\mathcal{E}\!\left(\mathrm{Spec}\!\left(\mathcal{M}_S,\;\mathcal{P}_{\mathcal{A}}(\mathcal{T};\,\mathcal{M}_T)\right)\right).
\end{equation}
Under this formulation, both $\mathcal{M}_T$ and $\mathcal{M}_S$ are fixed across tasks, enabling controlled analysis of the contribution of agent-driven data curation to student model specialization.

\subsection{Task Protocol}
\paragraph{\textbf{Task Input}}
As shown in Figure \ref{fig:main}(a), for each task the agent is provided with:
(1) a brief introduction of the evaluation setting;
(2) a basic overview of the target dataset, including dataset description, submission format, optional seed pool, and the public test set for validation;
(3) a fixed budget of teacher model API calls that the agent can use to synthesize data; and
(4) a fixed student model for domain specialization, together with corresponding standardized fine-tuning \& inference parameters.\footnote{By default, we adopt \texttt{Qwen3-30B-A3B} as the teacher model and \texttt{LLaMA-3.1-8B-Instruct} as the student model.}

\paragraph{\textbf{Task Output}}
The agent is tasked to produce training data $\widehat{\mathcal{D}}$ as a \texttt{submission.json} file that conforms to the required format.
The submission must be produced by the agent's generated code, with all instances generated via teacher-model API calls rather than directly written into the file.

\paragraph{\textbf{Task Evaluation}}
We evaluate the agent by improving the end-to-end performance of the student model.
Specifically, the student model is fine-tuned on the submission data $\widehat{\mathcal{D}}$ and then evaluated on the hidden private set. 
The resulting private-set performance gain (Section~\ref{subsec:metric}) serves as a measure of the agent's end-to-end data engineering capability.

\paragraph{\textbf{Task Environment}}
Our running environment enforces fixed budgets on teacher-model API calls and wall-clock time, and provides standardized interfaces for teacher API calls, student model fine-tuning, and public set evaluation, as detailed in Appendix \ref{app:env}.
In this setting, the agent focuses solely on the data engineering task by implementing the data curation logic through code generation.

\subsection{Dataset Preparation}
\label{sec:dataset}
We collect QA reasoning tasks from three representative domains: \textit{Science}, \textit{Code}, and \textit{Finance}, evaluating how agents adapt and improve through autonomous data engineering within each domain. 

\paragraph{\textbf{Dataset Selection}}
We select task domains that satisfy: 
(i) \emph{specialized tasks} that are not adequately covered by general-purpose pretraining, where targeted specialization is essential to unlock the model's full potential; 
(ii) \emph{direct evaluation}, enabling deterministic rule-based scoring serving as environment feedback without execution environments or LLM judgment; 
and (iii) \emph{broad reasoning pattern} across representative domains. 
Based on these criteria, we adopt SciBench \citep{DBLP:conf/icml/WangHL0ZSLZS024}, LiveCodeBench \citep{DBLP:conf/iclr/JainHGLYZWSSS25} Test Output Prediction (LCB-TOP), and FinanceReasoning \citep{DBLP:conf/acl/TangEMHLYRLJHHL25} for final evaluation.

\paragraph{\textbf{Dataset Standardization}}
We derive task descriptions from their official documentation and redesign the original evaluation logic to be fully rule-based by removing subjective or LLM judgment components. 
In addition, we provide a standardized sample submission file for each task that defines the required format for generated training data.
Ultimately, we normalize each task as:
\begin{itemize}[leftmargin=*]
    \item \textbf{Dataset Description}: an overview of the dataset, component illustration, and data examples.  
    \item \textbf{Evaluation Script}: a script extracting answers from responses and computing dataset scores.
    \item \textbf{Seed Data}: standardized raw materials for domain specialization, where agent visibility depends on the experiment setting.
    \item \textbf{Public Test Set}: the visible data split for procedural feedback during iterative optimization.
    \item \textbf{Private Test Set}: the hidden data split reserved exclusively for final performance evaluation.
    \item \textbf{Sample Submission}: the required task-specific data generation format.
\end{itemize}

\paragraph{\textbf{Dataset Partition}}
For seed data construction, we fix a budget of 1{,}000 instances per task and ensure that all seeds contain only raw questions and associated context without reference answers (examples in Appendix \ref{app:seed_examples}). 
Specifically, for the \textit{\textbf{Science}} task, we filter SciInstruct \cite{DBLP:conf/nips/ZhangHZDYWYD024} to retain instances with deterministic numeric answers, and then apply data selection strategies for quality.
For the \textit{\textbf{Code}} task, we draw seeds from LiveCodeBench releases v1--v6 via stratified sampling, further augmented with stratified samples from TACO \citep{DBLP:journals/corr/abs-2312-14852}. 
For the \textit{\textbf{Finance}} task, due to limited related resources, we sample half of FinanceReasoning as seed data. 
We then construct the public and private splits from SciBench, LCB-TOP, and the remaining portion of FinanceReasoning. 
The resulting Public Test Set and Private Test Set follow a 1:3 split ratio.
Throughout seed construction and test-set splitting, we enforce strict \textbf{stratified sampling} and rigorously \textbf{ensure zero overlap in problems and contexts to prevent data leakage}.

\subsection{Automatic Data Engineering Agent}
We investigate agentic data engineering under two representative scenarios: a single-turn completion setting (\textbf{One-Shot}) and a closed-loop, self-optimizing setting (\textbf{Iterative Agent}), both illustrated in Figure~\ref{fig:main} (b).

\paragraph{\textbf{One-Shot}}

In this setting, the agent generates the final submission in a single pass.
We provide the agent with a comprehensive prompt with the necessary task input.
The agent then drafts a strategy plan, implements it via \texttt{code.py}, and produces \texttt{submission.json} (Figure~\ref{fig:main}(b-ii)).
We allow up to 8 independent attempts to mitigate generation failure.
Once a valid submission is generated, the process terminates, and the submission is used to fine-tune the student model.

\paragraph{\textbf{Iterative Agent}}

In this setting, the agent is tasked with continuously enhancing model performance through a closed-loop data engineering process.
Inspired by recent advances in self-improving agents~\citep{DBLP:conf/nips/MadaanTGHGW0DPY23, DBLP:journals/corr/abs-2502-13138}, we investigate whether LLMs can apply such capabilities to data engineering by leveraging environmental feedback signals.
To this end, we design the Iterative Agent, as illustrated in Figure \ref{fig:main}(b-i), incorporating four operations:
\begin{itemize}[leftmargin=*]
  \item \textbf{Draft}. Guided by the task settings and the dataset description, the agent formulates a new data synthesis strategy plan by outlining a plan and implementing it via executable code. 
  \item \textbf{Debug}. When the generated code throws an error during execution, the agent analyzes the traceback to diagnose and fix errors, ensuring the script executes successfully.
  \item \textbf{Repair}. When the code executes successfully but the generated \texttt{submission.json} fails validation, the agent either refines the synthesis strategy to regenerate data or post-processes existing instances in the raw data, ensuring the submission meets the required quantity and format.
  \item \textbf{Improve}. Leveraging environmental feedback, the agent employs iterative improvement: it applies a \textit{greedy strategy} to select the solution with the highest public score from iteration history, consisting of the plan, code, and submission data, and optimizes it to evolve the synthesis strategy and enhance data quality.
\end{itemize}

Specifically, the process initiates with the \textit{\textbf{Draft}} operation.
The generated code for data curation first undergoes an execution check, and any failure triggers the \textit{\textbf{Debug}} operation. 
Upon successful execution, if the output fails the submission validation check (i.e., < 1{,}000 samples remain after format filtering), the process shifts to the \textit{\textbf{Repair}} operation.
We cap \textit{\textbf{Debug}} and \textit{\textbf{Repair}} operations at 3 consecutive attempts, restarting from \textit{\textbf{Draft}} if this limit is exceeded.
If the data validation check passes, the agent submits the curated data, receives feedback from the environment, and proceeds to the \textit{\textbf{Improve}} operation accordingly.
This iterative process enables the agent to simultaneously optimize synthesis strategies, prompt designs, and data distributions, continually driving student model specialization (see Appendix~\ref{appendix:improve_example} for a running example).

%% file: section/experiments.tex
\input{table/main_res}
\section{Experiments}

\subsection{Metric Definition}
\label{subsec:metric}
We assess the agentic data engineering capability in a training-based setting, where the student model's post-training performance gain directly reflects the agent's effectiveness.

\paragraph{\textbf{Relative Performance Gain (\%)}}
To enable consistent comparison across tasks, we report the \emph{relative performance gain} of the student model:
\begin{equation}
\mathrm{Gain}(\%) \;=\; \frac{\mathrm{Score}(\mathcal{M}_S^{\star}) - \mathrm{Score}(\mathcal{M}_S)}{\mathrm{Score}(\mathcal{M}_S)} \times 100
\end{equation}
where $\mathcal{M}_S$ denotes the initial student model and $\mathcal{M}_S^{\star}$ denotes the specialized student model fine-tuned on the agent's final data submission.
Positive values indicate successful model specialization, whereas negative values indicate performance degradation.
We follow each source benchmark's official evaluation metric, with all tasks evaluated by accuracy.
We also report the absolute accuracy of each run in Table~\ref{tab:raw_main} as a complementary view.

\paragraph{\textbf{Mean Attempts to Success (MATS)}}
MATS measures the average number of trial attempts to obtain a \emph{successful} data submission. 
Given a run with $N$ attempts, an attempt is marked successful if it generates a \texttt{submission.json} and the filtered submission retains at least $1{,}000$ instances after format validation filtering. 
We report
\begin{equation}
\mathrm{MATS} = \frac{N}{\sum_{i=1}^{N}\mathbb{I}\left[\mathrm{succ}(i)\right]}
\end{equation}
where $\mathrm{succ}(i)=1$ if the $i$-th attempt yields a successful submission and $\mathrm{succ}(i)=0$ otherwise.

\subsection{Experiment Setup}
\paragraph{\textbf{Execution Details}} We run experiments under the following budgets: 50{,}000 total teacher API calls per task ($\leq$5{,}000 per attempt), 3-hour limit per code execution (i.e., data synthesis), and 12-hour timeout limit per run, terminating once any budget is exhausted. 
For the \textit{\textbf{One-Shot}} scenario, we allow up to 8 attempts; \textit{\textbf{Iterative Agents}} run for at most 30 iterations. 
We fine-tune Llama-3.1-8B-Instruct as student model on $2\times$ H100 GPUs and deploy Qwen3-30B-A3B as the teacher model on $2\times$ H100 GPUs via vLLM with max concurrency of 80 (details in Appendix~\ref{app:exp_settings}).
To verify generalization, we also evaluate alternative teacher-student configurations (see Appendix~\ref{app:teacher_student}).
Each complete iteration cycle (synthesis $\to$ training $\to$ evaluation) takes 1--2 hours under this setting.
We conduct two independent runs and report the final mean performance gain (raw accuracy scores in Table~\ref{tab:raw_main}).

\paragraph{\textbf{Data Initialization Settings}} We evaluate both agents under two distinct settings: (1) \textbf{\textit{From Scratch}}: The agent must synthesize the entire dataset relying solely on the task description and teacher model API. (2) \textbf{\textit{With Seed}}: The agent is additionally provided with a seed pool of 1,000 raw questions (as described in Section \ref{sec:dataset}) to guide the data synthesis and exploration process.

\begin{figure*}[!t]
  \centering
  \includegraphics[width=\linewidth]{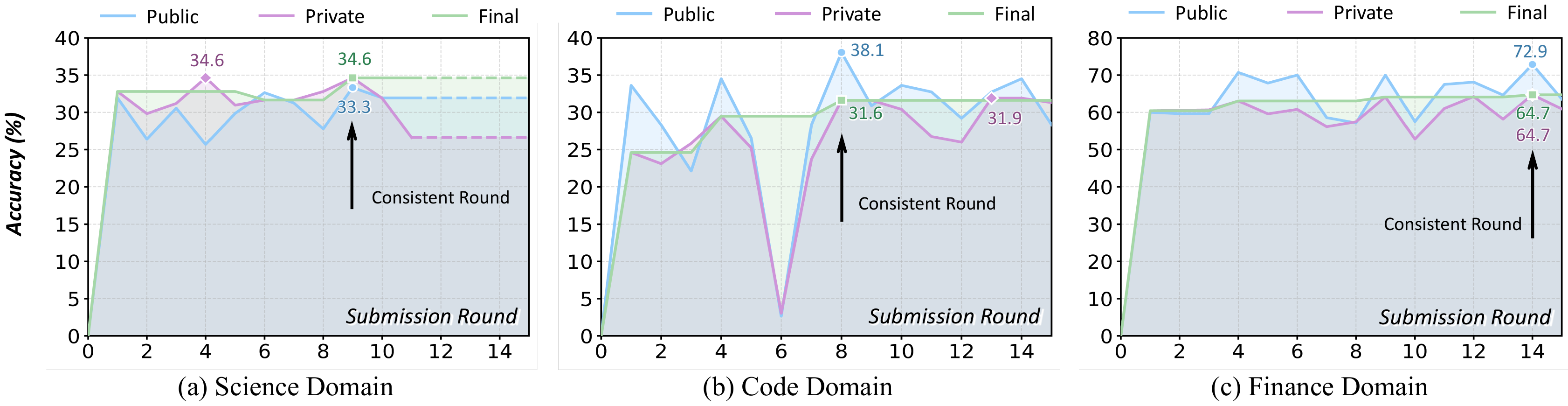}
  \vspace{-8mm}
  \caption{Iteration analysis of performance across successful submissions produced by the Iterative Agent.}
  \label{fig:iteration}
  \vspace{-3mm}
\end{figure*}

\begin{figure}[!t]
  \centering
  \includegraphics[width=\linewidth]{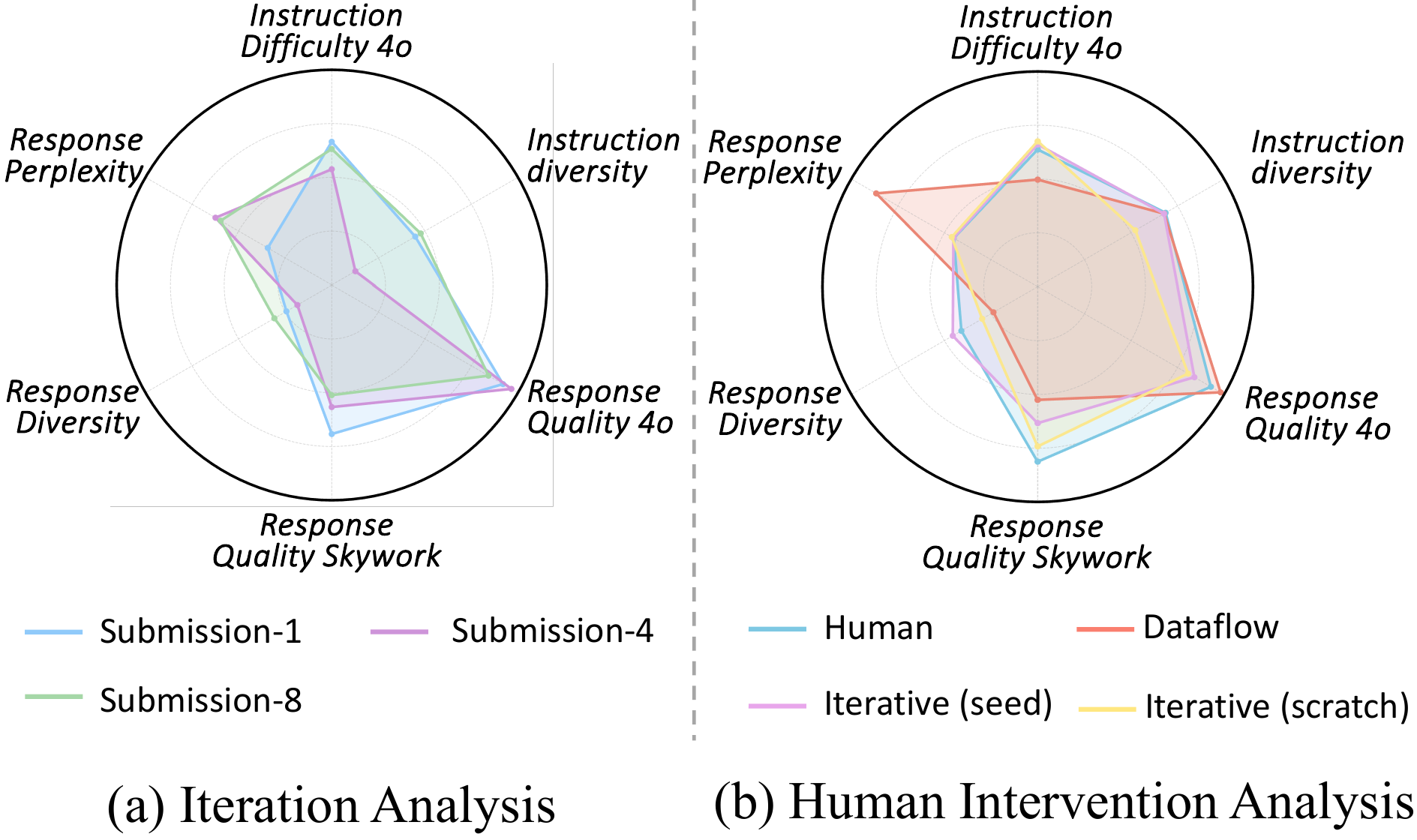}
  \vspace{-8mm}
  \caption{Quality evaluation of synthesized instructions.}
  \label{fig:instruction}
  \vspace{-6mm}
\end{figure}

\subsection{Main Results}

\paragraph{\textbf{Iterative optimization drives gains, while seed data ensures stability.}}
As shown in Table~\ref{tab:main_results}, \emph{Iterative Agents} consistently outperform \emph{One-Shot Agents}.
Specifically, in the \emph{from scratch} regime, GPT-5.2 improves its average relative gain from \textbf{40.73\%} to \textbf{57.29\%}, demonstrating the efficacy of LLMs in leveraging environment feedback for self-improvement.
Compared with \emph{One-Shot} generation, where a single error can corrupt the process (DeepSeek-V3.1's -4.58\% drop on Code), iterative mechanisms mitigate this by repeatedly improving exploration toward higher-quality solutions. 
Furthermore, adding a 1k seed pool consistently improves performance in both settings. 
This effect is strongest in the more fragile \emph{One-Shot from scratch} scenario, where most models obtain \textbf{30\%+} additional relative gains after seeds are introduced.
These results suggest that seed data serves to broaden the agent's coverage of the target task distribution while injecting essential domain-relevant knowledge, thereby reducing off-target generation and low-quality instances.

\paragraph{\textbf{Frontier LLMs have emerged as independent data engineers for end-to-end model specialization.}}
Even in the most fragile \emph{One-Shot from scratch} setting, most agents still deliver positive average gains, and GPT~5.2 attains roughly a \textbf{40\%} improvement over the base model. 
Without external knowledge, environment feedback, or any human-designed workflow, the submission must be produced by agent-written code independently.
Under these constraints, the observed gains provide concrete evidence of non-trivial \textbf{\textit{data engineering ability}}. The LLM agents can autonomously infer what supervision the model lacks, synthesize task-aligned instances, and curate a training set that generalizes to the hidden private task distribution.

\paragraph{\textbf{Compared to stronger models, weaker models benefit more from sophisticated agent frameworks to unlock capabilities.}}
Table~\ref{tab:main_results} reveals a consistent interaction between base model capability and the effectiveness of complex agent frameworks. 
Weaker models experience substantially larger improvements from these advanced designs: DeepSeek-V3.1 surges from \textbf{12.50\%} in the \emph{one-shot from scratch} baseline to \textbf{57.65\%} with iterative optimization and seed data, while stronger models such as GPT-5.2 and the Claude family show relatively modest gains under the same conditions. 
This pattern suggests that feedback-driven iterative optimization and seed data injection serve as critical guide rails for weaker models.

%% file: table/main_res.tex
\definecolor{lightgray}{gray}{0.95}

\begin{table*}[t]
\centering
\vspace{-3mm}
\scriptsize
\setlength{\tabcolsep}{2.5pt}
\resizebox{\textwidth}{!}{%
\begin{tabular}{l|cc|cc|cc|cc||cc|cc|cc|cc}
\toprule
\multirow{3}{*}{\textbf{Agent Models}}
  & \multicolumn{8}{c||}{\textbf{Specialization from Scratch}}
  & \multicolumn{8}{c}{\textbf{Specialization with Seed}} \\
\cmidrule(lr){2-9}\cmidrule(lr){10-17}
  & \multicolumn{2}{c|}{\textbf{Science}}
  & \multicolumn{2}{c|}{\textbf{Code}}
  & \multicolumn{2}{c|}{\textbf{Finance}}
  & \multicolumn{2}{c||}{\textbf{Avg.}}
  & \multicolumn{2}{c|}{\textbf{Science}}
  & \multicolumn{2}{c|}{\textbf{Code}}
  & \multicolumn{2}{c|}{\textbf{Finance}}
  & \multicolumn{2}{c}{\textbf{Avg.}} \\
\cmidrule(lr){2-3}\cmidrule(lr){4-5}\cmidrule(lr){6-7}\cmidrule(lr){8-9}
\cmidrule(lr){10-11}\cmidrule(lr){12-13}\cmidrule(lr){14-15}\cmidrule(lr){16-17}
  & \textbf{MATS} & \textbf{Gain}
  & \textbf{MATS} & \textbf{Gain}
  & \textbf{MATS} & \textbf{Gain}
  & \textbf{MATS} & \textbf{Gain}
  & \textbf{MATS} & \textbf{Gain}
  & \textbf{MATS} & \textbf{Gain}
  & \textbf{MATS} & \textbf{Gain}
  & \textbf{MATS} & \textbf{Gain} \\
\midrule
\rowcolor{lightgray}
\multicolumn{17}{c}{\bfseries\itshape One-Shot} \\
\midrule
GPT-5.2         & 2.0 & 35.66 & 1.5 & \underline{34.89} & 1.5 & \underline{51.63} & 1.67 & \textbf{\underline{40.73}} & 1.0 & 58.24 & 1.5 & \underline{52.12} & 1.5 & 13.72 & 1.33 & 41.36 \\
Qwen3-Max       & 1.0 & 37.69 & 1.5 & 32.03 & 3.5 & 30.00 & 2.00 & 33.24 & 2.0 & 49.34 & 2.0 & 32.77 & 1.5 & 59.39 & 1.83 & 47.17 \\
DeepSeek-R1     & 2.5 & \underline{47.94} & 2.0 & 4.75 & 2.0 & -6.72 & 2.17 & 15.32 & 1.0 & 83.60 & 2.0 & 0.45 & 2.0 & 53.72 & 1.67 & 45.92 \\
DeepSeek-V3.1   & 2.5 & 18.04 & 2.0 & -4.58 & 3.0 & 24.03 & 2.50 & 12.50 & 2.5 & 52.78 & 1.0 & 19.83 & 2.5 & 59.25 & 2.00 & 43.95 \\
Gemini-2.5-Pro  & 2.0 & 35.63 & 1.0 & 22.00 & 2.0 & 11.34 & 1.67 & 22.99 & 1.0 & \underline{87.01} & 1.0 & 20.56 & 1.0 & \underline{66.35} & 1.00 & \textbf{\underline{57.97}} \\
Claude-4-Sonnet & 3.0 & 30.17 & 2.0 & 15.53 & 1.5 & 17.46 & 2.17 & 21.05 & 2.5 & 79.48 & 1.0 & 14.07 & 2.0 & 61.34 & 1.83 & 51.63 \\
\midrule
\rowcolor{lightgray}
\multicolumn{17}{c}{\bfseries\itshape Iterative Agent} \\
\midrule
GPT-5.2         & 1.53 & \underline{70.58} & 1.75 & \underline{50.68} & 1.25 & \underline{50.60} & 1.51 & \textbf{\underline{57.29}} & 3.00 & 82.23 & 2.07 & 49.24 & 1.36 & 36.56 & 2.14 & 56.01 \\
Qwen3-Max       & 1.58 & 58.24 & 1.83 & 34.16 & 1.41 & 39.84 & 1.61 & 44.08 & 1.75 & 74.67 & 2.58 & 20.56 & 1.23 & 61.33 & 1.85 & 52.19 \\
DeepSeek-R1     & 2.13 & 37.01 & 3.63 & 11.92 & 1.56 & 26.56 & 2.44 & 25.16 & 2.75 & \underline{89.76} & 4.00 & 19.10 & 1.65 & 54.92 & 2.80 & 54.59 \\
DeepSeek-V3.1   & 7.13 & 45.91 & 2.17 & 24.10 & 1.80 & 40.28 & 3.70 & 36.76 & 1.47 & 78.14 & 1.24 & 29.15 & 1.31 & 65.66 & 1.34 & 57.65 \\
Gemini-2.5-Pro  & 1.60 & 41.79 & 1.71 & 16.24 & 1.13 & 35.06 & 1.48 & 31.03 & 1.63 & 46.62 & 1.55 & 37.77 & 2.16 & 64.11 & 1.78 & 49.50 \\
Claude-4-Sonnet & 2.61 & 58.93 & 1.74 & 22.71 & 2.35 & 39.69 & 2.23 & 40.44 & 2.17 & 82.08 & 2.24 & \underline{69.33} & 3.59 & \underline{68.36} & 2.67 & \textbf{\underline{73.26}} \\
\bottomrule
\end{tabular}%
}
\caption{\textbf{Main Results.} We report \textbf{MATS} (Mean Attempts to Successful Submission; lower is better) and relative performance \textbf{Gain} (\%) over the base \texttt{Llama-3.1-8B-Instruct} model (higher is better), using \texttt{Qwen3-30B-A3B} as the unified teacher model. Results are averaged over two runs, with the raw accuracy scores reported in Table~\ref{tab:raw_main}.}
\label{tab:main_results}
\end{table*}

%% file: section/analysis.tex
\section{Further Analysis}
\input{table/human_data}

\subsection{Iterative Data Optimization Analysis}
We conduct a controlled analysis to investigate how \textit{\textbf{Iterative Agents}} improve and what specific aspects are optimized during iteration.
Specifically, we increase the iteration and API-call limits and extend the time budget to 48 hours (Figure~\ref{fig:iteration}), with all runs generated using \texttt{GPT-5.2}.
With \emph{public} score already recorded during the execution loop, we re-evaluate the student model on the \emph{private} test set for every \emph{successful} submission. 
We also report a \emph{final} score which represents the private score of the best-performing submission on the public leaderboard up to time $t$, reflecting what the agent would actually select with the greedy strategy.

\paragraph{\textbf{Iterative Agent demonstrates steady overall improvement across iterations.}}
Figure~\ref{fig:iteration} shows that \emph{public}, \emph{private}, and \emph{final} scores exhibit a clear upward trend despite some fluctuations across iterations. 
Substantial gains typically emerge within the first 8 to 15 iterations, beyond which performance plateaus, indicating diminishing returns as the agent reaches the boundaries of its data awareness and cognitive capacity~\citep{DBLP:conf/nips/ShinnCGNY23}.

\paragraph{\textbf{Greedy public-score selection ensures robustness performance.}}
The fluctuations in Figure~\ref{fig:iteration} reflect the intrinsic variance of synthetic data curation, where minor changes of prompts or generation pipelines can substantially shift answer correctness and the data distribution. 
For instance, in Figure~\ref{fig:iteration} (b) (Round~6), a pipeline mistake causes a sharp drop in both public and private performance. 
The greedy selection rule mitigates such failures by retaining the best historical submission based on \emph{public} score~\citep{chen2021evaluatinglargelanguagemodels}. 
Given that \emph{public} and \emph{private} performance are largely aligned across iterations, this strategy yields a \emph{final} curve that is noticeably more stable, thus less susceptible to occasional catastrophic regressions.

\paragraph{\textbf{Iteration primarily drives improvements in data diversity.}}
Following prior work~\citep{DBLP:conf/acl/KimSY0LWGLWN25}, we conduct quality analysis on the generated submission data using six intrinsic metrics: instruction difficulty (GPT-4o assessment), instruction diversity (embedding similarity), response quality (GPT-4o assessment \& Skywork reward model), response diversity (embedding similarity), and response perplexity (LLaMA-3.1-8B).
The diagnostics reveal that both instruction and response diversity consistently increase over iterations, while response quality improves only marginally (example in Appendix~\ref{appendix:improve_example}).
Consistent with prior work~\citep{yu2024diversifyconquerdiversitycentricdata}, this indicates that iterations primarily expand and diversify synthesized questions rather than enhance quality for existing items.

\subsection{Human Involvement Influence Analysis}
To systematically examine how varying degrees of human involvement influence data curation, we study several settings as follows:

\begin{itemize}[leftmargin=*]
  \item \textbf{Human}. 
  We build a 2k training set by sampling from SciInstruct with output-length filtering and diversity-aware clustering~\citep{DBLP:journals/corr/abs-2506-04178}. We then use LLMs to rewrite instruction pairs into the target evaluation format. In this setting, both the data source and the synthesis pipeline are fully specified by humans.
  \item \textbf{DataFlow}. DataFlow provides a general synthesis from scratch pipeline with predefined strategies for generation, filtering, and refinement. We adopt it as a strong method representing a human-designed synthesis pipeline without relying on an external data source.
  \item \textbf{Iterative Agent (with seed / from scratch)}. We report the best-performing submission across all iterative rounds to approximate the current upper bound of the LLM data engineer without any human-designed strategies or recipes.
  
\end{itemize}

\paragraph{\textbf{Fully autonomous data engineering shows potential to outperform human-involved methods through iterative optimization.}}
Under the same constraint, the pipeline designed by GPT-5.2 surpasses the human-designed DataFlow framework (Table \ref{tab:human_data}).
First, frontier LLMs can flexibly adapt their pipeline design strategy to the target task, automatically aligning the synthesized data to the appropriate domain, difficulty, and output format, rather than relying on rigid human-designed logic, which is consistent with OPRO~\citep{yang2024largelanguagemodelsoptimizers}. 
Second, environmental feedback acts as a closed-loop signal that mimics the self-reflection process~\citep{DBLP:conf/nips/ShinnCGNY23}, enabling LLMs to continuously improve their data curation strategies and progressively shape the data distribution toward the specialized domain.

\paragraph{\textbf{LLMs can match human-level data complexity, while falling short in generating diverse data.}}
As illustrated in Figure~\ref{fig:instruction}(b), the \textit{from-scratch} agent successfully approaches the human baseline in \textit{Instruction Difficulty}, demonstrating its strong capability to self-curate challenging data. 
Nevertheless, it exhibits a performance drop in \textit{Instruction Diversity} and \textit{Response Diversity} compared to human-involved settings. 
This result reveals a basic limit of purely LLM-driven data engineering: the generated examples are high-quality but too repetitive.

\subsection{Failure Mode Analysis}
\label{sec:error_analysis}

\begin{figure*}[!t]
  \centering
  \includegraphics[width=\linewidth]{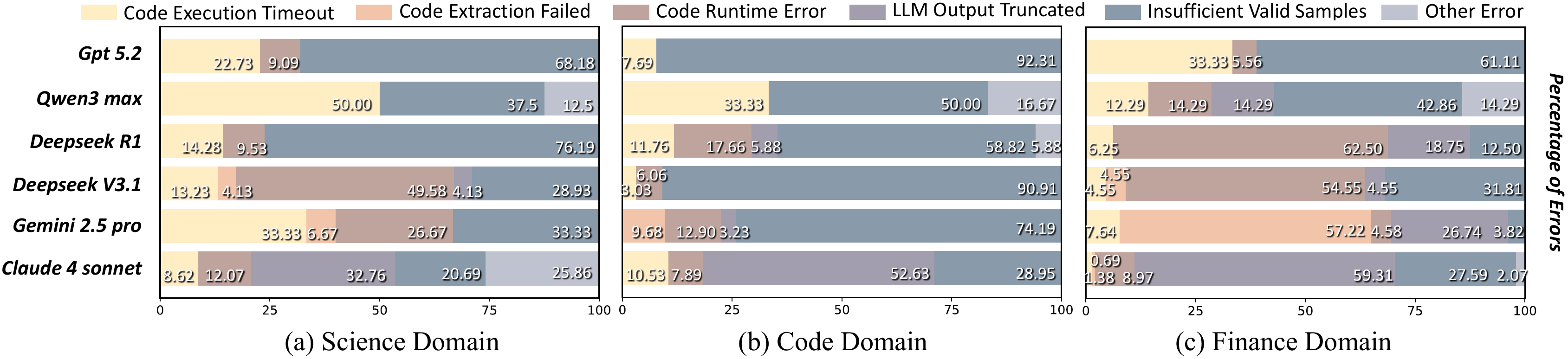}
  \vspace{-6mm}
  \caption{Error type analysis of valid submission generation failure.}
  \label{fig:error}
  \vspace{-4mm}
\end{figure*}

\paragraph{\textbf{Data Submission Failure}}

As shown in Table \ref{tab:main_results}, LLMs often fail to generate valid data in a single round. 
We perform detailed analysis based on the error-type breakdown (Figure~\ref{fig:error}). 

\begin{itemize}[leftmargin=*]
  \item \textbf{Lack of Quantity Assurance Awareness}. 
  As shown in Figure~\ref{fig:error}, \textit{Insufficient Valid Samples} dominates errors across most models (e.g., GPT-5.2 and DeepSeek-R1). While agents aggressively filter generated data, they lack the awareness to validate the final dataset size and dynamically replenish discarded samples, ultimately failing the 1,000-instance quantity check.
  
  \item \textbf{Weak Format Handling in Complex Domains}. 
  Error distribution is domain-dependent. \textit{Insufficient Valid Samples} is generally milder in text-based \textit{Finance} tasks. Conversely, \textit{Science} (requiring LaTeX) and \textit{Code} (requiring executable logic) impose formatting constraints, causing massive data rejection and extraction failures.
  
  \item \textbf{Potential Over-Engineering Trap}. 
  We observe an anomalously high rate of \textit{LLM Output Truncated} specifically for Claude-4-Sonnet (e.g., 52.63\% in Code and 59.31\% in Finance), revealing a tendency to design overly complex, verbose curation pipelines that exceed task requirements.
\end{itemize}

\paragraph{\textbf{Model Specialization Failure}}

Agents demonstrate capabilities for data engineering in most cases, but also encounter typical failure instances.

\begin{itemize}[leftmargin=*]
  \item \textbf{Distribution Shift of Data.} 
  In a \textit{from-scratch Science} failure case (Appendix~\ref{appendix:failure_cases_Science}), the agent hard-coded logic forcing 50\% of the data budget into just five narrow topics (e.g., Boltzmann distribution). 
  Instead of broad scientific sampling, this skewed generation caused a severe distribution shift. 
  Consequently, it induced catastrophic forgetting~\citep{luo2025empirical}, causing the student model to overfit to these specific sub-domains while losing broader competency.
  
  \item \textbf{Naive Rule-Based Augmentation.} 
    In a \textit{from-seed Code} failure case (Appendix~\ref{appendix:failure_cases_Code}), the agent employs a naive regex strategy to indiscriminately perturb numerical values, ignoring their distinct roles in control flow. 
    This severed the semantic link between instructions and executable logic, directly violating the SECON principle~\citep{10.1145/3686151}. 
    Consequently, instead of valid data expansion, the agent injected syntactically broken noise, severely degrading the student model's performance.
\end{itemize}

Overall, despite demonstrated capability to guide end-to-end data curation, \textbf{LLMs still lack robust post-generation safeguards for stringent quality assurance and reliable quantity control}.

%% file: table/human_data.tex
\begin{table}[!t]
\centering
\scriptsize
\setlength{\tabcolsep}{3pt}
\resizebox{\columnwidth}{!}{
\begin{tabular}{lllll}
\toprule
\textbf{Method} & \textbf{Source} & \textbf{Pipeline} & \textbf{Teacher} & \textbf{Gain} \\
\midrule
Human               & SciInstruct & Human   & None       & 84.95 \\
DataFlow            & None        & Human   & Qwen3-30B  & 65.82 \\
Iterative (seed)    & SciInstruct & GPT-5.2 & Qwen3-30B  & 93.19 \\
Iterative (scratch) & None        & GPT-5.2 & Qwen3-30B  & 76.76 \\
\bottomrule
\end{tabular}
}
\vspace{-3.5mm}
\caption{Comparison with human-involved settings. Detailed configurations see Appendix~\ref{app:human_involvement}.}
\label{tab:human_data}
\vspace{-5.5mm}
\end{table}

%% file: section/related_work.tex
\section{Related Work}

\noindent\textbf{LLM Agent.}
LLM agents~\citep{DBLP:journals/fcsc/WangMFZYZCTCLZWW24, DBLP:journals/chinaf/XiCGHDHZWJZZFWXZWJZLYDW25} 
leverage the reasoning capabilities of foundation 
models~\citep{DBLP:journals/corr/abs-2505-03418}, integrated with external 
tools~\citep{DBLP:conf/nips/SchickDDRLHZCS23} and environmental 
feedback~\citep{DBLP:conf/iclr/YaoZYDSN023} to complete tasks.
This revolution has catalyzed the emergence of specialized agents 
across diverse domains, ranging from autonomous data 
analysis~\citep{DBLP:journals/corr/abs-2306-07209} to scientific 
discovery~\citep{DBLP:journals/nature/BoikoMKG23}, with growing 
attention to agent memory~\citep{chen2026memprivacy} and 
reusable skill acquisition~\citep{DBLP:journals/corr/abs-2603-01145, zhang2026mmskills}.
A particularly relevant line of work investigates how agents 
continuously evolve through accumulated 
experience~\citep{DBLP:journals/corr/abs-2511-06449, ouyang2026skillos, DBLP:journals/corr/abs-2601-07055}, constructing structured 
knowledge from past successes and failures to improve future 
performance.
While these efforts focus on agents that \emph{consume} data to 
complete tasks, our work studies agents that \emph{produce and 
optimize} data as primary objective.

\noindent\textbf{Data Centric AI.}
In the early stages of LLM development, training heavily relied on 
high-quality human-annotated data~\citep{DBLP:conf/nips/StiennonO0ZLVRA20}. 
With the rapid depletion of naturally occurring human text, a crisis 
of data scarcity has become 
imminent~\citep{DBLP:conf/icml/VillalobosHSBHH24}. 
Pioneering work such as 
Self-Instruct~\citep{DBLP:conf/acl/WangKMLSKH23} introduced a 
paradigm shift, demonstrating that LLMs can synthesize training data 
from a small seed set to optimize 
themselves~\citep{taori_alpaca_2023, DBLP:journals/corr/abs-2306-11644}. 
More recent studies further improve the automated generation of 
synthetic data~\citep{DBLP:conf/iclr/Huang0GCZ00X00025, liang2025dataflow, DBLP:journals/corr/abs-2512-00884}, 
yet remain fundamentally dependent on data synthesis pipelines or 
recipes designed by humans.
Meanwhile, the search-verify-feedback 
paradigm~\citep{DBLP:journals/corr/abs-2411-11504} highlights the 
importance of automated verification as a supervision signal for 
post-training.
Autodata~\citep{kulikov2026autodata} further introduces a 
multi-agent system with predefined roles that iteratively generates 
challenging examples, though the architecture and quality criteria 
remain human-specified.
Concurrent to our work, 
PostTrainBench~\citep{DBLP:journals/corr/abs-2603-08640} benchmarks 
whether LLM agents can automate the full post-training pipeline with 
unconstrained choice over training methods, hyperparameters, and 
data sources.
DataPrep-Bench~\citep{liang2026dataprepbench} compares diverse data 
construction methods and quality metrics under a unified 
downstream-grounded protocol.
In contrast, our work fixes the training pipeline, parameter settings, teacher model, and student model, isolating data curation as the sole variable to 
study whether individual LLM agents can iteratively optimize 
training data driven by the downstream student model feedback.

%% file: section/conclusion.tex
\section{Conclusion}
We present a systematic analysis of \emph{Agentic Data Engineering
for Model Specialization} by requiring LLM agents to conduct end-to-end data engineering in a closed loop. 
Our results across \emph{Science}, \emph{Code}, and \emph{Finance} show that iterative agents consistently yield stronger and more stable specialization, with feedback-driven iteration improving both data strategy and alignment. 
In particular, GPT-5.2 reaches a 57.29\% average gain, demonstrating that LLM agents can autonomously author data curricula that drive substantial student model specialization. 
Our further failure analysis of the observed dominance of invalid submissions and failed specializations exposes the lack of data assurance awareness of current LLMs. 

%% file: appendix/dataset.tex
\section{Dataset Details}
\label{app:dataset_details}

To systematically analyze the capability of LLM agents in end-to-end data engineering, we curate datasets across three specialized domains: \textit{Science}, \textit{Code}, and \textit{Finance}. We elaborate on the source datasets and the rationale behind our construction choices below; the standardization protocol and partition rules follow Section~\ref{sec:dataset}.

\paragraph{Science.}
We build the Science task upon \textbf{SciBench}~\citep{DBLP:conf/icml/WangHL0ZSLZS024} and \textbf{SciInstruct}~\citep{DBLP:conf/nips/ZhangHZDYWYD024}. SciBench evaluates college-level scientific reasoning across physics, chemistry, and mathematics, providing a rigorous testbed for model specialization, while SciInstruct serves as a diverse instruction-tuning corpus suitable for seed construction. Since our environment requires deterministic rule-based scoring, we filter SciInstruct to retain instances with definitive numeric answers before applying quality-aware selection to form the seed pool. SciBench is used exclusively for evaluation, with zero overlap in problems and contexts against the seed pool.

\paragraph{Code.}
For the programming domain, we adopt \textbf{LiveCodeBench (LCB)}~\citep{DBLP:conf/iclr/JainHGLYZWSSS25} and \textbf{TACO}~\citep{DBLP:journals/corr/abs-2312-14852}. We focus on the Test Output Prediction sub-task of LCB (LCB-TOP), which requires predicting program execution outputs and thus demands deep algorithmic understanding. Seed instances are drawn from LCB releases v1--v6 via stratified sampling, augmented with stratified samples from TACO to broaden diversity. The evaluation sets are constructed exclusively from LCB-TOP under the same zero-overlap constraint.

\paragraph{Finance.}
For the financial domain, we adopt \textbf{FinanceReasoning}~\citep{DBLP:conf/acl/TangEMHLYRLJHHL25}, which targets deep financial logic, numerical reasoning, and domain-specific text comprehension. As high-quality open-source resources for complex financial reasoning are scarce, both the seed and the evaluation sets are derived from FinanceReasoning via disjoint stratified splits, ensuring no contextual leakage between the seed pool and the test sets.

%% file: appendix/exp_details.tex
\section{Main Result Details}
\label{app:res_details}

\input{table/base_model}

Table~\ref{tab:base_scores} details the original performance of our unified student model, \texttt{LLaMA-3.1-8B-Instruct}, across the three target domains. These scores reflect the model's capabilities on the private test sets prior to any instruction tuning or distillation. Specifically, the base model achieves accuracy scores of 16.74 on science domain (SciBench), 21.18 on code domain (LCB-TOP), and 39.93 on finance domain (FinanceReasoning). These results establish a performance baseline, highlighting the challenges of these specialized tasks for the off-the-shelf model and serving as a reference for quantifying the improvements gained through our synthetic data generation methods.

Table~\ref{tab:raw_main} provides the comprehensive raw data from our main experiments. We report two key metrics: MATS (Mean Attempts to Success Submission), which quantifies the efficiency of the agent in generating valid datasets, and Accuracy (\%), which measures the performance of the student model fine-tuned on the synthesized data. The results cover both \textit{Specialization from Scratch} and \textit{Specialization with Seed} settings across all three domains. To demonstrate the stability of our approach, we report results from at least two independent runs for each agent configuration.

\input{table/raw_main_exp}

%% file: table/base_model.tex
\begin{table}[!htbp]
\centering
\footnotesize
\setlength{\tabcolsep}{3pt}
\renewcommand{\arraystretch}{1.1}
\resizebox{\columnwidth}{!}{%
\begin{tabular}{@{}lllc@{}}
\toprule
\textbf{Task} & \textbf{Benchmark} & \textbf{Metric} & \textbf{Pri. Score} \\
\midrule
Science & SciBench         & Acc. (\%) & 16.74 \\
Code    & LCB-TOP          & Acc. (\%) & 21.18 \\
Finance & FinanceReasoning & Acc. (\%) & 39.93 \\
\bottomrule
\end{tabular}%
}
\caption{Base-model performance on each task. We use \texttt{LLaMA-3.1-8B-Instruct} as the unified base model.}
\label{tab:base_scores}
\end{table}

%% file: table/raw_main_exp.tex
\definecolor{lightgray}{gray}{0.95}

\begin{table*}[t]
\centering
\small
\setlength{\tabcolsep}{3pt}
\renewcommand{\arraystretch}{1.05}
\resizebox{\textwidth}{!}{%
\begin{tabular}{l|c|cc|cc|cc||cc|cc|cc}
\toprule
\multirow{3}{*}{\textbf{Agent Models}}
  & \multirow{3}{*}{\textbf{Runs}}
  & \multicolumn{6}{c||}{\textbf{Specialization from Scratch}}
  & \multicolumn{6}{c}{\textbf{Specialization with Seed}} \\
\cmidrule(lr){3-8}\cmidrule(lr){9-14}
  & 
  & \multicolumn{2}{c|}{\textbf{Science}}
  & \multicolumn{2}{c|}{\textbf{Code}}
  & \multicolumn{2}{c||}{\textbf{Finance}}
  & \multicolumn{2}{c|}{\textbf{Science}}
  & \multicolumn{2}{c|}{\textbf{Code}}
  & \multicolumn{2}{c}{\textbf{Finance}} \\
\cmidrule(lr){3-4}\cmidrule(lr){5-6}\cmidrule(lr){7-8}
\cmidrule(lr){9-10}\cmidrule(lr){11-12}\cmidrule(lr){13-14}
  & 
  & \textbf{MATS} & \textbf{Acc(\%)}
  & \textbf{MATS} & \textbf{Acc(\%)}
  & \textbf{MATS} & \textbf{Acc(\%)}
  & \textbf{MATS} & \textbf{Acc(\%)}
  & \textbf{MATS} & \textbf{Acc(\%)}
  & \textbf{MATS} & \textbf{Acc(\%)} \\
\midrule
\rowcolor{lightgray}
\multicolumn{14}{c}{\bfseries\itshape One-Shot} \\
\midrule
\multirow{2}{*}{GPT-5.2}
 & run1 & 1.00 & 15.83 & 2.00 & 27.96 & 1.00 & 59.83 & 1.00 & 30.73 & 2.00 & 32.22 & 2.00 & 46.60 \\
 & run2 & 2.00 & 29.59 & 1.00 & 29.18 & 2.00 & 61.26 & 1.00 & 22.25 & 1.00 & 32.22 & 1.00 & 44.22 \\
\multirow{2}{*}{Qwen3-Max}
 & run1 & 1.00 & 27.29 & 1.00 & 27.05 & 1.00 & 45.89 & 1.00 & 28.67 & 3.00 & 27.36 & 2.00 & 62.81 \\
 & run2 & 1.00 & 18.81 & 2.00 & 28.88 & 6.00 & 57.93 & 3.00 & 21.33 & 1.00 & 28.88 & 1.00 & 64.48 \\
\multirow{2}{*}{DeepSeek-R1}
 & run1 & 3.00 & 27.06 & 2.00 & 20.97 & 2.00 & 46.60 & 1.00 & 31.65 & 2.00 & 30.09 & 3.00 & 57.09 \\
 & run2 & 2.00 & 22.47 & 2.00 & 23.40 & 2.00 & 27.89 & 1.00 & 29.82 & 2.00 & 12.46 & 1.00 & 65.67 \\
\multirow{2}{*}{DeepSeek-V3.1}
 & run1 & 2.00 & 30.05 & 1.00 & 29.79 & 4.00 & 65.20 & 3.00 & 17.66 & 3.00 & 18.54 & 3.00 & 54.59 \\
 & run2 & 3.00 & 21.10 & 1.00 & 20.97 & 1.00 & 61.98 & 2.00 & 21.86 & 1.00 & 21.88 & 3.00 & 44.46 \\
\multirow{2}{*}{Gemini-2.5-Pro}
 & run1 & 2.00 & 17.66 & 1.00 & 25.23 & 2.00 & 42.07 & 2.00 & 30.05 & 1.00 & 29.79 & 4.00 & 65.20 \\
 & run2 & 2.00 & 27.75 & 1.00 & 26.44 & 2.00 & 46.84 & 3.00 & 21.10 & 1.00 & 20.97 & 1.00 & 61.98 \\
\multirow{2}{*}{Claude-4-Sonnet}
 & run1 & 3.00 & 20.64 & 2.00 & 23.71 & 2.00 & 46.01 & 4.00 & 28.67 & 1.00 & 37.08 & 2.00 & 60.55 \\
 & run2 & 3.00 & 22.94 & 2.00 & 25.23 & 1.00 & 47.79 & 1.00 & 31.42 & 1.00 & 11.24 & 2.00 & 68.30 \\
\midrule
\rowcolor{lightgray}
\multicolumn{14}{c}{\bfseries\itshape Iterative Agent} \\
\midrule
\multirow{2}{*}{GPT-5.2}
 & run1 & 1.80 & 25.46 & 1.00 & 30.09 & 1.17 & 60.79 & 3.50 & 32.34 & 1.80 & 31.00 & 1.60 & 48.75 \\
 & run2 & 1.25 & 31.65 & 2.50 & 33.74 & 1.33 & 59.48 & 2.50 & 28.67 & 2.33 & 32.22 & 1.11 & 60.31 \\
\multirow{2}{*}{Qwen3-Max}
 & run1 & 1.17 & 28.21 & 1.00 & 23.40 & 1.06 & 53.40 & 1.50 & 31.19 & 3.50 & 33.74 & 1.33 & 63.17 \\
 & run2 & 2.00 & 24.77 & 2.67 & 33.43 & 1.75 & 58.28 & 2.00 & 27.29 & 1.67 & 17.33 & 1.13 & 65.67 \\
\multirow{2}{*}{DeepSeek-R1}
 & run1 & 2.00 & 25.69 & 3.25 & 25.23 & 1.40 & 49.82 & 3.00 & 32.11 & 1.00 & 30.39 & 2.22 & 58.05 \\
 & run2 & 2.25 & 20.18 & 4.00 & 22.18 & 1.73 & 51.25 & 2.50 & 31.42 & 7.00 & 20.06 & 1.07 & 65.67 \\
\multirow{2}{*}{DeepSeek-V3.1}
 & run1 & 1.25 & 17.89 & 2.33 & 29.78 & 1.80 & 61.26 & 1.15 & 26.15 & 1.14 & 24.92 & 1.05 & 66.98 \\
 & run2 & 13.00 & 30.96 & 2.00 & 22.79 & 1.80 & 50.77 & 1.80 & 33.49 & 1.33 & 29.79 & 1.57 & 65.32 \\
\multirow{2}{*}{Gemini-2.5-Pro}
 & run1 & 1.60 & 19.72 & 1.71 & 22.49 & 1.06 & 46.48 & 1.75 & 33.03 & 1.50 & 32.22 & 1.92 & 66.70 \\
 & run2 & 1.60 & 27.75 & 1.71 & 26.74 & 1.20 & 61.38 & 1.50 & 16.06 & 1.60 & 26.14 & 2.40 & 64.36 \\
\multirow{2}{*}{Claude-4-Sonnet}
 & run1 & 4.00 & 26.15 & 1.27 & 26.14 & 3.14 & 50.77 & 3.33 & 28.44 & 1.14 & 39.21 & 3.17 & 66.15 \\
 & run2 & 1.21 & 27.06 & 2.20 & 25.84 & 1.56 & 60.79 & 1.00 & 32.52 & 3.33 & 32.52 & 4.00 & 68.30 \\
\bottomrule
\end{tabular}%
}
\caption{Raw scores of the main experiment. We report \textbf{MATS} (Mean Attempts to Success Submission) and the Accuracy (\textbf{Acc}, \%) of the fine-tuned student model \texttt{LLaMA-3.1-8B-Instruct}. Scores of the original student model are reported in Table~\ref{tab:base_scores}. The teacher model is set to \texttt{Qwen3-30B-A3B} globally.}
\label{tab:raw_main}
\end{table*}

%% file: appendix/exp_settings.tex
\section{Experiment Configuration Details}
\label{app:exp_settings}

In this section, we provide a comprehensive breakdown of the experimental configurations used in our study. This includes the hyperparameter settings for our data synthesis agents (both One-Shot and Iterative), the specific configurations for model training via LoRA, and the inference parameters employed for both the teacher and student models.

\subsection{One-Shot Agent Configuration}
The One-Shot Agent represents a baseline approach where the synthetic dataset is generated in a single pass without iterative refinement. 
The configuration is designed to balance generation speed with robustness against potential execution failures.

\begin{itemize}
    \item \textbf{Resource Constraints}: We limit the total runtime to 12 hours (Max Time Hours) and the dataset size to 2,000 samples (Dataset Size). We enforce a 3-hour limit per code execution (data synthesis) and a strict budget of 50,000 total teacher API calls per task ($\le$ 5,000 per attempt).
    \item \textbf{Teacher Model}: We utilize Qwen3-30B-A3B as the teacher model. To maximize throughput during the data generation phase, we set the Api Concurrency to 80, allowing parallel processing of multiple data points via the vLLM deployment.
    \item \textbf{Student Model}: The student model serves as the validator. It is hosted locally using vLLM with Vllm Max Num Seqs set to 128 to optimize inference throughput during the validation phase.
    \item \textbf{Robustness Mechanism}: We set Max Generation Attempts to 8. This allows the agent to retry the generation process up to 8 times if the code fails to execute or produces invalid JSON output.
\end{itemize}

\begin{tcolorbox}[title={Detailed Configuration for the One-Shot Agent.}, label={lst:one_shot_yaml}]
\begin{verbatim}
# General Settings
common:
  DATASET_SIZE: 2000
  MAX_TIME_HOURS: 12
  EXECUTION_TIMEOUT_MIN: 180
  MAX_GENERATION_ATTEMPTS: 8

# Teacher Model Settings (Generator)
teacher:
  TEACHER_MODEL: Qwen3-30B-A3B
  TOTAL_API_LIMIT: 50000
  SESSION_API_LIMIT: 5000
  API_CONCURRENCY: 80

# Student Model Settings (Validator)
student:
  LOCAL_MODEL: Llama-3_1-8B-Instruct
  VLLM_PORT: 8099
  VLLM_MAX_NUM_SEQS: 128
  VLLM_CONCURRENCY: 64
  VLLM_MAX_TOKENS: 8192
\end{verbatim}
\end{tcolorbox}

\subsection{Iterative Agent Configuration}
The Iterative Agent introduces a feedback loop where the agent analyzes the performance of the trained model and refines the dataset accordingly. This requires a more sophisticated configuration to handle the cycle of drafting, training, evaluating, and improving.

Key distinctions in the configuration include:
\begin{itemize}
    \item \textbf{Self-Correction Loop}: The agent runs for at most 30 iterations. Within each iteration, the agent may encounter errors in its proposed code. We define Max Debug Attempts as 3 for self-correction within a single cycle.
    \item \textbf{Resource Allocation}: The system adheres to the global 12-hour time limit and monitors the Total Api Limit of 50,000 calls to decide whether to continue refining or finalize the submission. Each complete iteration cycle takes approximately 1--2 hours.
\end{itemize}

The specific parameters are detailed in Listing~\ref{lst:iterative_yaml}.

\begin{tcolorbox}[title={Detailed Configuration for the Iterative Agent.}, label={lst:iterative_yaml}]
\begin{verbatim}
# General Settings
common:
  DATASET_SIZE: 2000
  MAX_TIME_HOURS: 12
  MAX_ITERATIONS: 30
  MAX_DEBUG_ATTEMPTS: 3

# Teacher Model (Data Generator)
teacher:
  TEACHER_MODEL: Qwen3-30B-A3B
  API_CONCURRENCY: 80

# Agent Core (Controller)
iterative-agent:
  AGENT_MODEL: your_agent_model
  env_vars:
    <<: [
    *common_settings,
    *teacher_config, 
    *student_config
    ]
\end{verbatim}
\end{tcolorbox}

\subsection{Model Training Parameters}
All synthetic datasets are evaluated by training a student model using Supervised Fine-Tuning (SFT). We adopt a parameter-efficient approach using Low-Rank Adaptation (LoRA) on 2$\times$ H100 GPUs.

The complete training configuration is provided in Listing~\ref{lst:model_train_yaml}.
 And the training configuration is standardized as follows:
\begin{itemize}
    \item \textbf{Optimization Strategy}: We use the AdamW optimizer with a learning rate of $1.0 \times 10^{-4}$ and a cosine learning rate scheduler. A warmup ratio of 0.1 is applied to stabilize the early training phase.
    \item \textbf{LoRA Configuration}: We target all linear layers (Lora Target: all) with a rank of 8 (Lora Rank) and an alpha of 16 (Lora Alpha). This configuration provides a good balance between adaptation capacity and parameter efficiency.
    \item \textbf{Compute Efficiency}: To accommodate hardware constraints, we set the per-device batch size to 1 but employ Gradient Accumulation Steps of 8. Training is performed in Bf16 precision to reduce memory usage. The model is trained for 3 epochs with evaluations performed every 500 steps.
\end{itemize}

\begin{tcolorbox}[title={Hyperparameters for Model Training (SFT with LoRA).}, label={lst:model_train_yaml}]
\begin{verbatim}
# Training Stage
stage: sft
finetuning_type: lora
lora_target: all
lora_rank: 8
lora_alpha: 16
lora_dropout: 0

# Hyperparameters
learning_rate: 1.0e-4
num_train_epochs: 3
lr_scheduler_type: cosine
warmup_ratio: 0.1
bf16: true

# Batch Size & Gradient Accumulation
per_device_train_batch_size: 1
gradient_accumulation_steps: 8
cutoff_len: 2048
\end{verbatim}
\end{tcolorbox}

\subsection{Model Inference Parameters}
The inference process involves two distinct phases: data generation (Teacher) and model evaluation (Student). The parameters for each phase are optimized for their specific objectives. Furthermore, we detail the model names and API versions of all evaluated agent models in the experiments.

\subsubsection{Teacher Model Inference}
The teacher model (Qwen3-30B-A3B) is deployed on separate 2$\times$ H100 GPUs via vLLM. The inference parameters are managed dynamically by the agent code:
\begin{itemize}
    \item \textbf{Concurrency}: We utilize an asyncio.Semaphore with a limit of 80 (Api Concurrency) to maximize generation throughput against the vLLM server.
    \item \textbf{Sampling}: The teacher model uses standard sampling parameters to ensure diversity in the generated synthetic data.
\end{itemize}

\subsubsection{Student Model Inference (Evaluation)}
The student model is deployed locally using the vLLM engine. The configuration focuses on memory stability and evaluation throughput:
\begin{itemize}
    \item \textbf{System Configuration}: We explicitly set Gpu Memory Utilization to 0.85 to reserve GPU memory for activation overheads, preventing Out-Of-Memory (OOM) errors during long-sequence processing. The Max Num Seqs is set to 128 to fully utilize the GPU's parallel processing capacity.
    \item \textbf{Tensor Parallelism}: The system automatically detects the number of available GPUs and scales the Tensor Parallel Size accordingly.
    \item \textbf{Dynamic Sampling}: During evaluation, sampling parameters (e.g., temperature) are not hardcoded but are generated dynamically based on the specific dataset requirements (defined in evaluate.py).
\end{itemize}

These settings are summarized in Listing~\ref{lst:model_infer_yaml}.

\begin{tcolorbox}[title={Inference Parameters for Teacher and Student Models.}, label={lst:model_infer_yaml}]

\begin{verbatim}
# Teacher Model
teacher_inference:
  model: Qwen3-30B-A3B
  deployment: vLLM (2x H100)
  concurrency_limit: 80
  temperature: default

# Student Model
student_system:
  engine: vLLM
  gpu_memory_utilization: 0.85
  max_model_len: 8192
  max_num_seqs: 128
  tensor_parallel_size: auto

student_sampling:
  temperature: dynamic
  top_p: dynamic
\end{verbatim}
\end{tcolorbox}

\subsubsection{Agent Model Details}
To facilitate experimental reproducibility and control of variables, the specific Agent Model API versions are listed below.
\begin{tcolorbox}[title={Exact API Model Versions.}]  
\begin{verbatim}  
GPT-5.2: 
    gpt-5.2-20251211  
Qwen3-Max: 
    qwen3-max-20250923  
DeepSeek-R1: 
    deepseek-r1-20250528  
DeepSeek-V3.1: 
    deepseek-v3.1-20250922  
Gemini-2.5-Pro: 
    gemini-2.5-pro-20250617  
Claude-4-Sonnet: 
    claude-sonnet-4-20250514  
\end{verbatim}  
\end{tcolorbox}  

\subsection{Human Involvement Analysis Setup}
\label{app:human_involvement}

Across all settings, we fix the teacher model to Qwen3-30B-A3B. We cap the API budget at 5{,}000 calls for \textit{Iterative Agent}, and assign \textit{DataFlow} with official configuration of 6{,}000 calls. Each final synthesized training set contains 2{,}000 examples. The student model and training \& inference parameters are kept identical across all settings.

%% file: appendix/platform.tex
\section{Platform Design}
\label{app:env}
Our benchmark is accompanied by an execution platform that enables agents to synthesize data and run end-to-end model training within a controlled environment  (Figure~\ref{fig:main} (a)).

\paragraph{\textbf{Agent Toolkit}}
We provide a set of programmatic tools that agents can directly invoke, including \texttt{check\_submission} (format and schema validation), \texttt{train\_model} (fine-tuning on the submitted data), \texttt{evaluate\_dataset} (public set evaluation and bad case demonstration), and \texttt{api\_count} (API-usage tracking). 
We additionally ship a lightweight library with pre-defined helper functions for common operations during code generation, such as batched API calls and robust answer parsing. 
These tools allow an LLM agent to implement its own data-synthesis logic while seamlessly integrating with the full pipeline.

\paragraph{\textbf{Execution \& Evaluation Environment}}
We offer an isolated execution environment that separates both (i) the per-run workspace and (ii) the runtime execution context at the thread level, enabling safe concurrent runs while preventing data contamination. 
During execution, the platform continuously monitors API calls and remaining time, automatically terminating runs that exceed the prescribed budgets.

After each run completes, we replay the evaluation on the final \texttt{submission.json} by re-training the model on the submitted dataset and re-evaluating it on the held-out private test set, ensuring that the reported gains are attributable to the data.

%% file: appendix/student_teacher.tex
\section{Generalization Across Different Teacher-Student Configurations}
\label{app:teacher_student}

In our main experiments, we employ a fixed teacher-student model configuration. This serves as a controlled evaluation setup to provide a consistent scale for comparison, following common practices in prior empirical studies (e.g., using a fixed user simulator in $\tau^2$-Bench \cite{DBLP:journals/corr/abs-2506-07982}). 

To demonstrate that our findings are robust and not overfit to a specific model pair, we conduct additional experiments using the GPT-5.2 agent under varying teacher and student configurations on \textit{Science} task. Results summarized in Table~\ref{tab:teacher_student_ablation}.

\begin{table}[ht]
\centering
\resizebox{\columnwidth}{!}{
\begin{tabular}{lllc}
\toprule
\textbf{Agent} & \textbf{Teacher} & \textbf{Student} & \textbf{Gain (\%)} \\
\midrule
One-Shot  & Qwen3-30B-A3B & Qwen2.5-1.5B-Instruct & 51.89 \\
Iterative & Qwen3-30B-A3B & Qwen2.5-1.5B-Instruct & \textbf{59.43} \\
\midrule
One-Shot  & DeepSeek-V3.2 & LLaMA-3.1-8B-Instruct & 24.43 \\
Iterative & DeepSeek-V3.2 & LLaMA-3.1-8B-Instruct & \textbf{53.16} \\
\bottomrule
\end{tabular}
}
\caption{Performance gain of the GPT-5.2 agent across different teacher and student model configurations. The iterative approach consistently outperforms the one-shot baseline regardless of the underlying models.}
\label{tab:teacher_student_ablation}
\end{table}

As shown in Table~\ref{tab:teacher_student_ablation}, while the absolute value of the Gain (\%) naturally varies depending on the base capabilities of the chosen models, the overall trends remain highly consistent. Specifically, the iterative agent consistently yields positive gains and maintains a significant performance gap over the one-shot baseline across all configurations. 
These results confirm that the fixed teacher-student setting in our main text successfully serves its purpose of providing a consistent evaluation scale, and the effectiveness of autonomous data engineering generalizes well to other model combinations.

%% file: appendix/leakage_analysis.tex
\section{Data Leakage Analysis}
\label{app:data_leakage}

We examine whether the agent-generated training data directly reuse examples from the public or private evaluation sets. 
The private set is never exposed to either the agent or the teacher model.
The public-set samples are visible to the agent for task understanding and error analysis, while not directly provided to the teacher model for data synthesis.

We collect all submitted data across the three tasks and compare them against the corresponding public and private sets using exact match, fuzzy match with a threshold of $0.85$, and n-gram overlap with a threshold of $0.8$. 
We additionally measure internal redundancy among generated examples at both the input-only and input--output-pair levels.

As shown in Table~\ref{tab:data_leakage}, we find no direct overlap between the submitted data and either evaluation split under any of the three matching criteria. 
Although input-level repetition ranges from $3.06\%$ to $9.84\%$, redundancy among complete input--output pairs remains low, with a maximum of $1.49\%$.

\begin{table}[t]
    \centering
    \small
    \renewcommand{\arraystretch}{1.08}
    \resizebox{\columnwidth}{!}{%
    \begin{tabular}{llccc}
        \toprule
        \textbf{Comparison} & \textbf{Metric}
        & \textbf{Sci. (\%)} & \textbf{Code (\%)} & \textbf{Fin. (\%)} \\
        \midrule
        Public
        & Exact match       & $0.00$ & $0.00$ & $0.00$ \\
        & Fuzzy match       & $0.00$ & $0.00$ & $0.00$ \\
        & N-gram match      & $0.00$ & $0.00$ & $0.00$ \\
        \midrule
        Private
        & Exact match       & $0.00$ & $0.00$ & $0.00$ \\
        & Fuzzy match       & $0.00$ & $0.00$ & $0.00$ \\
        & N-gram match      & $0.00$ & $0.00$ & $0.00$ \\
        \midrule
        Redundancy
        & Input only        & $3.06$ & $7.69$ & $9.84$ \\
        & Input + output    & $0.58$ & $0.00$ & $1.49$ \\
        \bottomrule
    \end{tabular}%
    }
    \caption{Overlap between agent-submitted training data and the public and private evaluation sets, together with internal data redundancy reported.}
    \label{tab:data_leakage}
\end{table}

These results indicate that the LLM data engineers do not directly copy or closely reproduce examples from our evaluation splits according to the examined criteria. 
However, given the public availability of the benchmarks and the limited transparency of model pre-training corpora, potential benchmark exposure during pre-training remains difficult to avoid.

%% file: appendix/generalize.tex
\section{Model Generalization Analysis}
\label{app:generalization}

Model specialization aims to improve domain-specific performance, but may lead to degradation in general capabilities. 
Prior work has shown that full fine-tuning can cause catastrophic forgetting~\cite{DBLP:journals/corr/abs-2308-08747}, while LoRA may mitigate, but not completely eliminate, such forgetting~\cite{DBLP:journals/tmlr/BidermanPOPGJKH24}. 

To examine whether the data curated by autonomous data engineers introduce such degradation, we evaluate the final specialized downstream models on GSM8K and MMLU under both from-scratch and with-seed settings.
\begin{table}[t]
    \centering
    \small
    \renewcommand{\arraystretch}{1.08}
    \resizebox{\columnwidth}{!}{%
    \begin{tabular}{llcc}
        \toprule
        \textbf{Domain} & \textbf{Setting}
        & \textbf{GSM8K} & \textbf{MMLU} \\
        \midrule
        Base Model & Base
        & $0.7657 \pm 0.0117$ & $0.6839 \pm 0.0037$ \\
        \midrule
        SciBench & From scratch
        & $0.7801 \pm 0.0114$ & $0.6838 \pm 0.0037$ \\
        LCB-TOP & From scratch
        & $0.7823 \pm 0.0112$ & $0.6841 \pm 0.0037$ \\
        Finance & From scratch
        & $0.7665 \pm 0.0117$ & $0.6854 \pm 0.0037$ \\
        \midrule
        SciBench & With seed
        & $0.7908 \pm 0.0112$ & $0.6753 \pm 0.0037$ \\
        LCB-TOP & With seed
        & $0.7953 \pm 0.0111$ & $0.6771 \pm 0.0037$ \\
        Finance & With seed
        & $0.7824 \pm 0.0114$ & $0.6832 \pm 0.0037$ \\
        \bottomrule
    \end{tabular}%
    }
    \caption{General-capability performance of the base model and the downstream models specialized on different tasks.}
    \label{tab:generalization}
\end{table}

As shown in Table~\ref{tab:generalization}, GSM8K performance improves across all specialization settings, with gains ranging from $0.08$ to $2.96$ percentage points over the base model. 
This improvement suggests that the synthesized reasoning data may transfer to general mathematical reasoning. MMLU performance remains close to that of the base model, with slight improvements in some from-scratch settings and mild decreases in the with-seed settings.

Overall, the downstream models broadly retain their general capabilities after specialization. While these results do not constitute a comprehensive evaluation of catastrophic forgetting, they show no substantial degradation on GSM8K and MMLU under the evaluated settings.

%% file: appendix/improve_analysis.tex
\onecolumn
\clearpage
\newpage
\section{Running Example of Iterative Agent} 
\label{appendix:improve_example}
To clearly demonstrate how the \textit{Iterative Agent} optimizes its own data synthesis pipeline, we have selected an optimization example of \texttt{GPT-5.2} on the \textit{Science} task to illustrate the specific details of the iterative improvement.

Here is the original plan drafted by the model:

\begin{tcolorbox}[title=Original Plan of GPT-5.2 in Science Task.]
\textbf{1. Define the instruction format:}
\begin{itemize}
  \item Define a single fixed \texttt{"instruction"} string matching \texttt{sample\_submission.json} style
\end{itemize}

\textbf{2. Build topic templates spanning:}
\begin{itemize}
  \item \textbf{Mathematics:} calculus (integrals, series, multivariable), differential equations, linear algebra, probability/statistics
  \item \textbf{Physics:} mechanics, E\&M, circuits, waves/optics, thermodynamics
  \item \textbf{Chemistry:} gases (vdW/ideal), equilibrium, kinetics, thermodynamics, electrochemistry, colligative properties
\end{itemize}

\textbf{3. Generate parameterized prompts (per template) requiring the teacher to:}
\begin{itemize}
  \item Write a textbook-style problem with given numbers and a specified answer unit
  \item Provide a step-by-step solution including computations
  \item End with: \texttt{The answer is therefore \textbackslash boxed\{...\}.} where \texttt{...} is a decimal with exactly three digits
\end{itemize}

\textbf{4. Bulk generation:}
\begin{itemize}
  \item Use \texttt{api\_generate\_batch} with batch size 80 to produce $\sim$2200 candidates (buffer for filtering)
\end{itemize}

\textbf{5. Post-process and filter:}
\begin{itemize}
  \item Keep only items whose output contains at least one \texttt{\textbackslash boxed\{...\}} and whose last \texttt{\textbackslash boxed\{\}} matches a decimal number with 3 digits
  \item Ensure output ends with the exact final sentence
  \item Ensure fields \texttt{instruction}, \texttt{input}, \texttt{output} are non-empty strings
\end{itemize}

\textbf{6. Regenerate if needed:}
\begin{itemize}
  \item If $<2000$ valid samples, regenerate only the deficit with stricter formatting reminders
\end{itemize}

\textbf{7. Finalize dataset:}
\begin{itemize}
  \item Shuffle, truncate to first 2000, and write to \texttt{../submission/submission.json}
\end{itemize}

\textbf{8. Checkpointing:}
\begin{itemize}
  \item Save checkpoints every 200 valid samples to avoid data loss
\end{itemize}
\end{tcolorbox}
Based on self-reflection and environmental feedback, \texttt{GPT-5.2} proposed the optimization approach shown below based on the original solution:

\begin{tcolorbox}[title=Improvement Plan of GPT-5.2 in Science Task.]
\textbf{1. Improve solution quality by:}
\begin{itemize}
   \item Generating more focused, step-by-step solutions without excessive verbosity
   \item Ensuring all calculations are complete and accurate
   \item Requiring clear final answers in proper boxed format
\end{itemize}

\textbf{2. Enhance problem diversity by:}
\begin{itemize}
   \item Creating problems across broader difficulty ranges
   \item Including more applied/real-world scientific scenarios
   \item Balancing theoretical and computational problems
\end{itemize}

\textbf{3. Better prompt engineering:}
\begin{itemize}
   \item More specific instructions for concise, accurate solutions
   \item Explicit requirements for complete calculations
   \item Template-based solution structure to ensure consistency
\end{itemize}

\textbf{4. Quality control:}
\begin{itemize}
   \item Filter out incomplete or malformed solutions
   \item Validate that solutions have proper final answers
   \item Ensure mathematical notation is correct
\end{itemize}
\end{tcolorbox}
Guided by the improvement plan above, the model generated a more complex and robust code version that covers a broader scope and includes more challenging questions. 
The improved code corresponding to this optimization is shown below.

\begin{lstlisting}[caption={Correnponding Code to Improve Data Quality}, label={lst:code_improvement}]
import os
import re
import json
import random
import hashlib
from typing import List, Dict, Tuple
from services.library import api_generate_batch, get_available_api_calls

OUT_PATH = "../submission/submission.json"
API_BATCH = 80
TARGET_N = 2000
MAX_API_CALLS_THIS_ITER = 5000

INSTRUCTION = (
    "Please provide a clear and step-by-step solution for a scientific problem in the categories of Chemistry, Physics, or Mathematics. "
    "The problem will specify the unit of measurement, which should not be included in the answer. "
    "Express the final answer as a decimal number with three digits after the decimal point. "
    "Conclude the answer by stating \"The answer is therefore \\\\boxed{[ANSWER]}.\""
)

BOX_RE = re.compile(r"\\boxed\{([^}]*)\}")
THREE_DEC_RE = re.compile(r"^-?\d+\.\d{3}$")
UNIT_SENT_RE = re.compile(r"The unit of the answer is\s+[^.]+\.", re.IGNORECASE)

def extract_last_boxed(text: str):
    boxes = BOX_RE.findall(text or "")
    return boxes[-1].strip() if boxes else None

def required_final_sentence(boxed_value: str) -> str:
    return f"The answer is therefore \\\\boxed{{{boxed_value}}}."

def ends_with_required_sentence(text: str) -> bool:
    last = extract_last_boxed(text)
    if last is None:
        return False
    return text.rstrip().endswith(required_final_sentence(last))

def is_valid_item(item: Dict[str, str]) -> bool:
    if not isinstance(item, dict):
        return False
    for k in ("instruction", "input", "output"):
        if k not in item or not isinstance(item[k], str) or not item[k].strip():
            return False

    out = item["output"]
    last = extract_last_boxed(out)
    if last is None or not THREE_DEC_RE.match(last):
        return False
    if not ends_with_required_sentence(out):
        return False

    # Input should look like a question and include unit statement
    inp = item["input"]
    if "Question:" not in inp:
        return False
    if not UNIT_SENT_RE.search(inp):
        return False

    # Ensure boxed answer has no units (just numeric already enforced)
    return True

def normalize_text(s: str) -> str:
    return re.sub(r"\s+", " ", (s or "").strip())

def fingerprint_input(inp: str) -> str:
    return hashlib.md5(normalize_text(inp).encode("utf-8")).hexdigest()

def save_json(data: List[Dict[str, str]], path: str):
    os.makedirs(os.path.dirname(path), exist_ok=True)
    with open(path, "w", encoding="utf-8") as f:
        json.dump(data, f, ensure_ascii=False, indent=2)

def load_existing(path: str) -> List[Dict[str, str]]:
    if not os.path.exists(path):
        return []
    with open(path, "r", encoding="utf-8") as f:
        obj = json.load(f)
    return obj if isinstance(obj, list) else []

# ---------- Improved prompting ----------
TOPICS = [
    # Mathematics
    ("Mathematics", "Calculus", "definite integral (substitution/parts)"),
    ("Mathematics", "Calculus", "improper integral / convergence"),
    ("Mathematics", "Series", "Taylor/Maclaurin approximation"),
    ("Mathematics", "Multivariable", "double integral / change of variables"),
    ("Mathematics", "Optimization", "Lagrange multipliers"),
    ("Mathematics", "Differential Equations", "first-order linear IVP with application"),
    ("Mathematics", "Differential Equations", "second-order ODE (mass-spring/damping)"),
    ("Mathematics", "Linear Algebra", "eigenvalues/eigenvectors numeric"),
    ("Mathematics", "Probability", "expectation/variance continuous RV"),
    ("Mathematics", "Statistics", "MLE / confidence interval numeric"),
    # Physics
    ("Physics", "Mechanics", "work-energy with friction / incline / spring"),
    ("Physics", "Mechanics", "momentum/impulse collision"),
    ("Physics", "Mechanics", "circular motion/banked curve"),
    ("Physics", "Electricity & Magnetism", "electric field/potential superposition"),
    ("Physics", "Circuits", "RC transient time to reach a voltage"),
    ("Physics", "Circuits", "DC circuit (equivalent resistance/current)"),
    ("Physics", "Waves", "standing waves / beat frequency"),
    ("Physics", "Thermodynamics", "ideal gas process (W,Q, U)"),
    ("Physics", "Optics", "thin lens / mirror imaging"),
    # Chemistry
    ("Chemistry", "Thermodynamics", " G and equilibrium constant"),
    ("Chemistry", "Gases", "van der Waals / compression factor"),
    ("Chemistry", "Equilibrium", "buffer pH / K_a, K_b"),
    ("Chemistry", "Kinetics", "rate laws / half-life"),
    ("Chemistry", "Electrochemistry", "Nernst equation cell potential"),
    ("Chemistry", "Solutions", "colligative properties ( T_f/ T_b)"),
]

UNITS_BY_DOMAIN = {
    "Mathematics": ["unitless", "s", "m", "kg", "Pa", "J"],  # math may be unitless
    "Physics": ["m/s", "m/s^2", "N", "J", "W", "C", "V", "Hz", "K", "Pa"],
    "Chemistry": ["kJ/mol", "J/mol", "mol/L", "atm", "Pa", "K", "V", "s", "g/mol"],
}

def build_generation_prompt(seed: int) -> str:
    rnd = random.Random(seed)
    domain, area, style = rnd.choice(TOPICS)
    difficulty = rnd.choice(["introductory", "intermediate", "advanced"])
    scenario = rnd.choice([
        "applied real-world scenario",
        "textbook-style theoretical scenario",
        "lab measurement scenario with given uncertainties ignored",
        "engineering estimation scenario",
    ])
    unit = rnd.choice(UNITS_BY_DOMAIN[domain])

    # Encourage concise but complete structure
    return f"""
Create ONE SciBench-style example.

Hard requirements:
- Domain: {domain}. Area: {area}. Style: {style}. Difficulty: {difficulty}.
- Scenario flavor: {scenario}.
- Write the QUESTION first, starting with exactly: "Question: "
- The question must be self-contained and include all constants needed.
- The question must include exactly one sentence: "The unit of the answer is {unit}."
  (If {unit} is not appropriate for your question, replace it with an appropriate unit and keep the same sentence pattern.)
- Then write the SOLUTION with a clear structure:
  1) Given / Find
  2) Key equations
  3) Step-by-step calculation (show intermediate numeric values)
  4) Final line exactly: The answer is therefore \\boxed{{X.XXX}}.
- Output must be concise: aim for ~12-25 lines; no filler.
- Final boxed value must be a decimal with exactly three digits; no units in the box.
- Ensure the last \\boxed{{...}} in the output is the final answer.

Return JSON only:
{{"input": "...", "output": "..."}}
""".strip()

def build_repair_prompt(item: Dict[str, str]) -> str:
    # Ask teacher to keep question, rewrite solution concisely + correct format.
    inp = item.get("input", "")
    out = item.get("output", "")
    return f"""
You will REPAIR a dataset example for SciBench numeric answering.

Keep the QUESTION text exactly as-is.
Rewrite the SOLUTION to be correct, concise, and fully computed.

Hard constraints for the rewritten solution:
- Use a clear structure (Given/Find, equations, calculations).
- Do not be overly verbose (target ~12-25 lines).
- The final answer must be a decimal with exactly three digits after the decimal point.
- Do NOT include units in the boxed answer.
- The last line MUST be exactly: The answer is therefore \\boxed{{X.XXX}}.
- Ensure the last \\boxed{{...}} is the final numeric answer.

Return JSON only:
{{"input": "<same question>", "output": "<new solution>"}}

QUESTION:
{inp}

ORIGINAL SOLUTION (may be wrong/malformed):
{out}
""".strip()

def call_teacher_json(prompts: List[str]) -> List[Dict]:
    # Must call in fixed batch size 80
    assert len(prompts) == API_BATCH, f"api_generate_batch must be called with batch size {API_BATCH}"
    sys_prompt = "You are a meticulous STEM textbook solution writer. Follow constraints exactly and output valid JSON only."
    return api_generate_batch(prompts, sys_prompt=sys_prompt, json=True)

def parse_generated(objs: List) -> List[Dict[str, str]]:
    items = []
    for o in objs:
        if isinstance(o, dict) and "input" in o and "output" in o:
            items.append({"instruction": INSTRUCTION, "input": o["input"], "output": o["output"]})
    return items

def dedup_keep_order(items: List[Dict[str, str]]) -> List[Dict[str, str]]:
    seen = set()
    out = []
    for it in items:
        fp = fingerprint_input(it.get("input", ""))
        if fp in seen:
            continue
        seen.add(fp)
        out.append(it)
    return out

def main():
    random.seed(20260208)

    existing = load_existing(OUT_PATH)
    print(f"Loaded {len(existing)} raw items from {OUT_PATH}")

    # Separate valid and invalid
    valid = []
    invalid = []
    for it in existing:
        if is_valid_item(it):
            valid.append(it)
        else:
            # coerce to expected keys if partially missing
            if isinstance(it, dict) and "input" in it and "output" in it:
                invalid.append({"instruction": INSTRUCTION, "input": it.get("input", ""), "output": it.get("output", "")})

    valid = dedup_keep_order(valid)
    print(f"Valid after strict filtering + dedup: {len(valid)}; candidates for repair: {len(invalid)}")

    # Save immediate checkpoint of filtered-valid
    save_json(valid[:TARGET_N], OUT_PATH)

    api_calls_used = 0

    # ---- Repair pass (single pass over as many invalid as needed) ----
    # We'll repair up to the deficit and not exceed call budget.
    def deficit():
        return max(0, TARGET_N - len(valid))

    repair_pool = invalid[:]  # order as loaded
    repair_idx = 0
    while deficit() > 0 and repair_idx < len(repair_pool):
        if api_calls_used >= MAX_API_CALLS_THIS_ITER:
            break
        if get_available_api_calls() <= 0:
            break

        batch_items = repair_pool[repair_idx:repair_idx + API_BATCH]
        if len(batch_items) < API_BATCH:
            break  # keep strict batching; remaining can be handled by generation
        repair_idx += API_BATCH

        prompts = [build_repair_prompt(it) for it in batch_items]
        objs = call_teacher_json(prompts)
        api_calls_used += 1

        repaired = parse_generated(objs)
        repaired_valid = [it for it in repaired if is_valid_item(it)]
        if repaired_valid:
            valid.extend(repaired_valid)
            valid = dedup_keep_order(valid)

        if len(valid) % 200 < len(repaired_valid):
            save_json(valid[:TARGET_N], OUT_PATH)
            print(f"Checkpoint after repair: {min(len(valid), TARGET_N)}")

        print(f"Repair call {api_calls_used}: got {len(repaired)} parsed, {len(repaired_valid)} valid; total valid {len(valid)}")

    # ---- Top-up generation ----
    gen_seed_base = 900000
    gen_round = 0
    while deficit() > 0:
        if api_calls_used >= MAX_API_CALLS_THIS_ITER:
            break
        if get_available_api_calls() <= 0:
            break

        prompts = [build_generation_prompt(gen_seed_base + gen_round * API_BATCH + i) for i in range(API_BATCH)]
        objs = call_teacher_json(prompts)
        api_calls_used += 1
        gen_round += 1

        items = parse_generated(objs)
        items_valid = [it for it in items if is_valid_item(it)]
        if items_valid:
            valid.extend(items_valid)
            valid = dedup_keep_order(valid)

        if len(valid) % 200 < len(items_valid) or len(valid) >= TARGET_N:
            save_json(valid[:TARGET_N], OUT_PATH)
            print(f"Checkpoint after gen: {min(len(valid), TARGET_N)}")

        print(f"Gen call {api_calls_used}: parsed {len(items)}, valid {len(items_valid)}; total valid {len(valid)}")

        # If validity is unexpectedly low, we could tighten further, but keep runtime simple.

    # Finalize
    random.shuffle(valid)
    final = valid[:TARGET_N]
    save_json(final, OUT_PATH)
    print(f"Final saved: {len(final)} items to {OUT_PATH}. API calls used this iteration: {api_calls_used}")

if __name__ == "__main__":
    main()
\end{lstlisting}

%% file: appendix/failure_analysis.tex
\section{Code Snippets for Failure Analysis} 
\label{appendix:failure_cases}

In this section, we provide the specific code implementations generated by the agents that correspond to the failure cases discussed in Section~\ref{sec:error_analysis}.

\subsection{Science Task: Partial Distribution Shift.}
\label{appendix:failure_cases_Science}
Listing~\ref{lst:science_collapse} presents the \texttt{code.py} generated during the \textit{from-scratch} iterative Science task.

\begin{lstlisting}[caption={Agent Code That Induced Data Distribution Shift}, label={lst:science_collapse}]
import os
import re
import json
import math
import random
from typing import List, Dict, Any, Optional
from services.library import api_generate_batch, get_available_api_calls

BEST_PATH = "../submission/submission_best.json"
OUT_PATH = "../submission/submission.json"

INSTRUCTION = (
    "Please provide a clear and step-by-step solution for a scientific problem in the categories "
    "of Chemistry, Physics, or Mathematics. The problem will specify the unit of measurement, "
    "which should not be included in the answer. Express the final answer as a decimal number "
    "with three digits after the decimal point. Conclude the answer by stating "
    "\"The answer is therefore \\\\boxed{[ANSWER]}.\""
)

# Must be 65 prompts per call.
BATCH_SIZE = 65

# Accept one-or-more backslashes before boxed (handles '\boxed' and '\\boxed' in stored strings)
BOX_RE = re.compile(r"\\+boxed\{([^}]*)\}")
FINAL_LINE_RE = re.compile(r"The answer is therefore\s*\\+boxed\{([^}]*)\}\.\s*$")

NUM_RE = re.compile(
    r"^[\s]*"
    r"(?P<sign>[+-]?)"
    r"(?P<num>(?:\d+(?:\.\d*)?|\.\d+))"
    r"(?:[eE](?P<exp>[+-]?\d+))?"
    r"[\s]*$"
)

UNIT_CUE_RE = re.compile(
    r"(unit\s+of\s+the\s+answer\s+is|units?\s*:|answer\s+should\s+be\s+in|the\s+unit\s+is|\bin\s+\$?\\?[a-zA-Z]+)",
    re.IGNORECASE,
)

# Corruption / low-quality heuristics seen in existing samples.
BAD_TOKEN_RE = re.compile(
    r"(\\nrac|[^\\]rac\{|boxed\{\s*\}|@@|\?\?|nan|inf)", re.IGNORECASE
)

def load_json_list(path: str) -> List[Dict[str, Any]]:
    if not os.path.exists(path):
        return []
    try:
        with open(path, "r", encoding="utf-8") as f:
            data = json.load(f)
        return data if isinstance(data, list) else []
    except Exception:
        return []

def save(path: str, data: List[Dict[str, Any]]) -> None:
    os.makedirs(os.path.dirname(path), exist_ok=True)
    with open(path, "w", encoding="utf-8") as f:
        json.dump(data, f, ensure_ascii=False, indent=2)

def parse_number(s: str) -> Optional[float]:
    s = s.strip()
    if not NUM_RE.match(s):
        return None
    try:
        x = float(s)
        if math.isnan(x) or math.isinf(x):
            return None
        return x
    except Exception:
        return None

def normalize_final_line(output: str) -> Optional[str]:
    out = output.rstrip()
    boxes = BOX_RE.findall(out)
    if not boxes:
        return None
    last_box_raw = boxes[-1].strip()
    val = parse_number(last_box_raw)
    if val is None:
        return None
    val_3 = f"{val:.3f}"
    lines = out.splitlines()
    if not lines:
        return None
    lines[-1] = f"The answer is therefore \\boxed{{{val_3}}}."
    return "\n".join(lines)

def validate_entry(inp: Any, out: Any) -> bool:
    if not isinstance(inp, str) or not isinstance(out, str):
        return False
    inp = inp.strip()
    out = out.strip()

    if not inp.startswith("Question:"):
        return False
    if not UNIT_CUE_RE.search(inp):
        return False

    if BAD_TOKEN_RE.search(out):
        return False

    # Ensure final line matches required template
    if not FINAL_LINE_RE.search(out):
        return False

    # Ensure last boxed number is parseable
    boxes = BOX_RE.findall(out)
    if not boxes:
        return False
    if parse_number(boxes[-1]) is None:
        return False

    # Ensure solution is not trivial/answer-only: require at least 6 lines
    if len(out.splitlines()) < 6:
        return False

    return True

def sanitize_item(inp: Any, out: Any) -> Optional[Dict[str, str]]:
    if not isinstance(inp, str) or not isinstance(out, str):
        return None
    norm = normalize_final_line(out)
    if norm is None:
        return None
    inp_s = inp.strip()
    out_s = norm.strip()
    if not validate_entry(inp_s, out_s):
        return None
    return {"instruction": INSTRUCTION, "input": inp_s, "output": out_s}

SYS_PROMPT_TARGETED = (
    "You are generating instruction-tuning data for solving college-level scientific problems.\n"
    "Return ONLY valid JSON (no markdown, no extra text).\n"
    "Schema: {\"input\": string, \"output\": string}\n"
    "Hard rules:\n"
    "- input MUST start with exactly 'Question:'\n"
    "- input MUST explicitly state the unit of the final answer in a sentence: 'The unit of the answer is ... .'\n"
    "- output MUST be a correct step-by-step solution with unit conversions shown.\n"
    "- output MUST have at least 6 lines.\n"
    "- output MUST end with EXACT last line: The answer is therefore \\\\boxed{NUMBER}.\n"
    "- NUMBER must be numeric only (no units) and prefer decimal with three digits.\n"
    "Focus heavily on these topics (rotate among them):\n"
    "1) Two-level system / Boltzmann population ratios with energies in cm^-1; use hc/kB=1.4388 cm*K.\n"
    "2) Photoelectric/ionization energy: wavelength in nm to eV; subtract electron KE from v.\n"
    "3) Coriolis deflection for projectile at given latitude; use Earth rotation Omega=7.292e-5 s^-1.\n"
    "4) Manometer pressure conversions (cm H2O, mmHg) and computing R from PV=nRT; avoid factor-1000 errors.\n"
    "5) Orbital mechanics energy changes between circular orbits; use mu=3.986e14 m^3/s^2, Re=6.371e6 m.\n"
    "Use standard constants when needed: g=9.81, kB=1.381e-23 J/K, h=6.626e-34 J s, c=2.998e8 m/s, "
    "e=1.602e-19 C, 1 eV=1.602e-19 J.\n"
)

SYS_PROMPT_GENERAL = (
    "You are generating instruction-tuning data for solving college-level scientific problems.\n"
    "Return ONLY valid JSON (no markdown, no extra text).\n"
    "Schema: {\"input\": string, \"output\": string}\n"
    "Rules:\n"
    "- input MUST start with exactly 'Question:'\n"
    "- input MUST explicitly state the unit of the final answer in a sentence: 'The unit of the answer is ... .'\n"
    "- output MUST be a correct step-by-step solution.\n"
    "- output MUST have at least 6 lines.\n"
    "- output MUST end with EXACT last line: The answer is therefore \\\\boxed{NUMBER}.\n"
    "- NUMBER must be numeric only and prefer decimal with three digits.\n"
    "Cover a broad mix of undergraduate topics across calculus, ODEs, probability/statistics, mechanics, E&M, circuits, "
    "thermodynamics, equilibrium, kinetics, electrochemistry, optics.\n"
    "Use standard constants when needed: g=9.81, R=8.314, k=8.988e9, h=6.626e-34, c=2.998e8.\n"
)

def build_prompt(seed: int, mode: str) -> str:
    random.seed(seed)
    if mode == "targeted":
        focus = random.choice([
            "two-level Boltzmann population with energy separation in cm^-1",
            "photoelectric/ionization energy from wavelength and electron speed",
            "Coriolis deflection for a projectile fired due north/south at given latitude",
            "manometer pressure conversion and computing gas constant R from measurements",
            "orbital mechanics: energy required to move between circular orbits including synchronous orbit"
        ])
        return (
            "Generate ONE original, solvable, college-level quantitative problem and its correct step-by-step solution.\n"
            f"Topic MUST be: {focus}.\n"
            "Hard constraints:\n"
            "1) Return exactly one JSON object: {\"input\":..., \"output\":...}.\n"
            "2) input starts with 'Question:' and explicitly states the unit in: 'The unit of the answer is ... .'\n"
            "3) output shows all key equations and unit conversions and has at least 6 lines.\n"
            "4) output ends with EXACT last line: The answer is therefore \\\\boxed{NUMBER}.\n"
            "5) NUMBER is numeric only, no units, and prefer three decimals.\n"
            f"Seed tag: {seed}"
        )
    else:
        return (
            "Generate ONE original, solvable, college-level scientific problem (Math/Physics/Chemistry) "
            "and its correct step-by-step solution.\n"
            "Ensure multi-step reasoning and intermediate computations; avoid trivial one-liners.\n"
            "Hard constraints:\n"
            "1) Return exactly one JSON object: {\"input\":..., \"output\":...}.\n"
            "2) input starts with 'Question:' and explicitly states the unit in: 'The unit of the answer is ... .'\n"
            "3) output has at least 6 lines.\n"
            "4) output ends with EXACT last line: The answer is therefore \\\\boxed{NUMBER}.\n"
            "5) NUMBER is numeric only, no units, and prefer three decimals.\n"
            f"Seed tag: {seed}"
        )

def main(target_total: int = 2000, max_calls_cap: int = 5000) -> None:
    # 1) Start from best, filter hard for quality
    best = load_json_list(BEST_PATH)
    data: List[Dict[str, str]] = []
    seen_inputs = set()

    for d in best:
        if not isinstance(d, dict):
            continue
        item = sanitize_item(d.get("input"), d.get("output"))
        if item is None:
            continue
        if item["input"] in seen_inputs:
            continue
        data.append(item)
        seen_inputs.add(item["input"])

    save(OUT_PATH, data)

    # 2) Regenerate until we reach 2000
    if len(data) >= target_total:
        save(OUT_PATH, data[:target_total])
        return

    available_calls = get_available_api_calls()
    calls_budget = min(max_calls_cap, available_calls)

    seed_base = 20260129
    call_idx = 0

    while len(data) < target_total and call_idx < calls_budget:
        mode = "targeted" if (call_idx % 2 == 0) else "general"
        sys_prompt = SYS_PROMPT_TARGETED if mode == "targeted" else SYS_PROMPT_GENERAL

        prompts = [build_prompt(seed_base + call_idx * 1000 + i, mode=mode) for i in range(BATCH_SIZE)]
        responses = api_generate_batch(prompts, sys_prompt=sys_prompt, json=True)

        for resp in responses:
            if not isinstance(resp, dict):
                continue
            item = sanitize_item(resp.get("input"), resp.get("output"))
            if item is None:
                continue
            if item["input"] in seen_inputs:
                continue
            data.append(item)
            seen_inputs.add(item["input"])
            if len(data) >= target_total:
                break

        save(OUT_PATH, data)
        call_idx += 1

    save(OUT_PATH, data[:target_total])

if __name__ == "__main__":
    main(target_total=2000, max_calls_cap=5000)
\end{lstlisting}



\subsection{Code Task: Naive Rule-based Augmentation.}
\label{appendix:failure_cases_Code}
Listing~\ref{lst:code_failure} presents the \texttt{code.py} generated during the \textit{from-seed} iterative Code task.

\begin{lstlisting}[caption={Agent Code That Employed Naive Rule-based Augmentation}, label={lst:code_failure}]
import os
import json
import re
import random
import uuid
import argparse
import logging
import math
from typing import List, Dict, Any, Optional
from datetime import datetime
from collections import Counter

logging.basicConfig(
    level=logging.INFO,
    format='%(asctime)s - [DATA_PIPELINE] - %(levelname)s - %(message)s',
    datefmt='%Y-%m-%d %H:%M:%S'
)
logger = logging.getLogger(__name__)

DEFAULT_TARGET_SIZE = 5000
PERTURBATION_RANGE = (-5, 5)
MUTATION_PROBABILITY = 0.8 
MIN_VALUE_CLAMP = 1

NUMBER_PATTERN = re.compile(r'(?<![a-zA-Z_])\d+(?![a-zA-Z_])')

class DataAugmenter:
    def __init__(self, seed_path: str, output_path: str, target_size: int):
        self.seed_path = seed_path
        self.output_path = output_path
        self.target_size = target_size
        self.stats = Counter()
        self.generated_ids = set()

    def load_seeds(self) -> List[Dict[str, Any]]:

        if not os.path.exists(self.seed_path):
            logger.error(f"Seed file not found at {self.seed_path}")
            return []
        
        try:
            with open(self.seed_path, 'r', encoding='utf-8') as f:
                data = json.load(f)
            
            valid_seeds = [
                item for item in data 
                if all(k in item for k in ('instruction', 'input', 'output'))
            ]
            logger.info(f"Loaded {len(valid_seeds)} valid seeds from {len(data)} total entries.")
            return valid_seeds
        except json.JSONDecodeError as e:
            logger.critical(f"Failed to parse seed JSON: {e}")
            return []

    def _stochastic_perturbation(self, text: str) -> str:
        def replace_match(match):
            if random.random() > MUTATION_PROBABILITY:
                return match.group()

            try:
                original_val = int(match.group())
                
                shift = random.randint(*PERTURBATION_RANGE)
                
                new_val = max(MIN_VALUE_CLAMP, original_val + shift)
                
                return str(new_val)
            except ValueError:
                return match.group()

        return NUMBER_PATTERN.sub(replace_match, text)

    def generate_variant(self, seed_item: Dict[str, Any]) -> Optional[Dict[str, Any]]:
        try:
            new_input = self._stochastic_perturbation(seed_item['input'])
            
            new_output = self._stochastic_perturbation(seed_item['output'])
            
            variant = {
                "instruction": seed_item['instruction'],
                "input": new_input,
                "output": new_output,
                "meta": {
                    "origin": "augmented",
                    "parent_id": seed_item.get("id", "unknown"),
                    "aug_method": "regex_perturbation_v2"
                }
            }
            
            if new_input == seed_item['input'] and new_output == seed_item['output']:
                self.stats['skipped_no_change'] += 1
                return None
                
            return variant
            
        except Exception as e:
            logger.warning(f"Failed to generate variant: {e}")
            self.stats['errors'] += 1
            return None

    def validate_dataset(self, dataset: List[Dict[str, Any]]) -> List[Dict[str, Any]]:

        unique_data = []
        seen_hashes = set()
        
        for item in dataset:
            content_hash = hash(f"{item['input']}|{item['output']}")
            
            if content_hash in seen_hashes:
                self.stats['duplicates_removed'] += 1
                continue
                
            seen_hashes.add(content_hash)
            unique_data.append(item)
            
        return unique_data

    def run(self):
        start_time = datetime.now()
        logger.info(f"Starting pipeline at {start_time}")
        
        seeds = self.load_seeds()
        if not seeds:
            return

        dataset = []
        dataset.extend(seeds)
        logger.info(f"Initialized with {len(seeds)} seed samples.")

        pbar_interval = self.target_size // 10
        while len(dataset) < self.target_size:
            base_seed = random.choice(seeds)
            variant = self.generate_variant(base_seed)
            
            if variant:
                dataset.append(variant)
                self.stats['generated'] += 1
            
            current_count = len(dataset)
            if current_count % pbar_interval == 0 and current_count > 0:
                logger.info(f"Progress: {current_count}/{self.target_size} samples generated...")

        logger.info("Running final validation and deduplication...")
        final_dataset = self.validate_dataset(dataset)
        
        final_dataset = final_dataset[:self.target_size]
        
        self.save_data(final_dataset)
        
        duration = datetime.now() - start_time
        logger.info(f"Pipeline completed in {duration}.")
        logger.info(f"Statistics: {dict(self.stats)}")

    def save_data(self, data: List[Dict[str, Any]]):
        os.makedirs(os.path.dirname(self.output_path), exist_ok=True)
        with open(self.output_path, 'w', encoding='utf-8') as f:
            json.dump(data, f, indent=2, ensure_ascii=False)
        logger.info(f"Successfully saved {len(data)} entries to {self.output_path}")

def parse_args():
    parser = argparse.ArgumentParser(description="Rule-based Data Augmentation Tool")
    parser.add_argument('--seed_path', type=str, default='../data/seed/seed.json', help='Path to seed data')
    parser.add_argument('--output_path', type=str, default='../submission/submission.json', help='Path to save submission')
    parser.add_argument('--target_size', type=int, default=5000, help='Target dataset size')
    return parser.parse_args()

if __name__ == "__main__":
    args = parse_args()
    
    augmenter = DataAugmenter(
        seed_path=args.seed_path,
        output_path=args.output_path,
        target_size=args.target_size
    )
    
    try:
        augmenter.run()
    except KeyboardInterrupt:
        logger.info("Pipeline interrupted by user. Saving partial progress...")
        pass
    except Exception as e:
        logger.exception("Fatal pipeline error")
        exit(1)
\end{lstlisting}

%% file: appendix/seed_example.tex

\section{Seed Examples}
\label{app:seed_examples}

In this section, we present representative seed examples used in our experiments for the Science, Code, and Finance domains. These examples are extracted directly from the \texttt{seed.json} files of the respective datasets.

\subsection{Science Domain}
The seed data for Sci-Bench primarily consists of complex scientific problems involving physics and chemistry calculations.

\begin{lstlisting}
[
    {
      "question": "Three identical metal spheres have the same diameter. Spheres 1 and 2 carry equal like charges
       Q, with separation much greater than their diameter, and experience force F. Sphere 3 is uncharged with an
      insulating handle. If sphere 3 touches sphere 1, then touches sphere 2, and is removed, what is the new
      interaction force between spheres 1 and 2?"
    },
    {
      "question": "In an isolated town of 5000 inhabitants, the spread of an epidemic is such that the rate of
      spread is jointly proportional to the number of infected and uninfected people. If 160 people are infected at
      the start and 1200 are infected after one week, how long does it take for 80% of the population (4000 people)
     to become infected?"
    },
    ...
]
\end{lstlisting}

\subsection{Code Domain}
The seed data for the Code task includes algorithmic problems with problem descriptions, examples, and test inputs.

\begin{lstlisting}
[
    {
      "question_content": "Given n, a and d as the number of terms, first term and common difference respectively
      of an Arthimetic Series. Find the sum of the series upto nth term.\n \nExample 1:\nInput: 5 1 3\nOutput:
      35\nExplanation: Series upto 5th term is\n1 4 7 10 13, so sum will be 35.\nExample 2:\nInput: 3 1 2\n
      Output: 9\nExample: Series upto 3rd term is \n1 3 5, so sum will be 9.\n \nYour Task:\nYou don't need to
      read or print anything. Your task is to complete the function sum_of_ap() which takes n, a and d as input
        parameter and returns the sum of the series.\n \nExpected Time Complexity: O(1)\nExpected Space Complexity:
      O(1)\n \nConstranits:\n1 <= n, a, d <= 100",
      "test_input": "5 1 3"
    },
    {
      "question_content": "Let $f_{x} = c^{2x-6} \\cdot f_{x-1} \\cdot f_{x-2} \\cdot f_{x-3}$ for $x \\ge 4$.
      \n\nYou have given integers $n$, $f_{1}$, $f_{2}$, $f_{3}$, and $c$. Find $f_{n} \\bmod (10^{9}+7)$.
      \n\n\n-----Input-----\n\nThe only line contains five integers $n$, $f_{1}$, $f_{2}$, $f_{3}$, and $c$ ($4
      \\le n \\le 10^{18}$, $1 \\le f_{1}$, $f_{2}$, $f_{3}$, $c \\le 10^{9}$).\n\n\n-----Output-----\n\nPrint
      $f_{n} \\bmod (10^{9} + 7)$.\n\n\n-----Examples-----\nInput\n5 1 2 5 3\n\nOutput\n72900\n\nInput\n17 97 41
      37 11\n\nOutput\n317451037\n\n\n\n-----Note-----\n\nIn the first example, $f_{4} = 90$, $f_{5} = 72900$.
      \n\nIn the second example, $f_{17} \\approx 2.28 \\times 10^{29587}$.",
      "test_input": "5 1 2 5 3"
    },
    ...
]
\end{lstlisting}

\subsection{Finance Domain}
The Finance-Reasoning seed data comprises specific financial scenarios (context) and quantitative questions requiring reasoning over that context.
\begin{lstlisting}
[
    {
      "question": "What is Alice's new adjusted monthly mortgage payment after the fixed-rate period for the
      remaining 10 years? Answer in dollars, rounded to the nearest cent.",
      "context": "Alice took a 15-year fixed-rate mortgage with a principal amount of $250,000 at an annual
      interest rate of 4.5%. After the fixed-rate period ended, the remaining principal balance was $150,000.
        Her mortgage transitioned to an adjustable-rate with the current index rate at 2% and a bank margin of 1.5%.
      She wants to calculate her new monthly payment for the remaining 10 years of the mortgage under these new
      terms, assuming there are no rate caps."
    },
    {
      "question": "What is the difference in the high and low prices of the common stock in the fourth quarter of
      2019? Answer to two decimal places.",
      "context": "{\"2019: -- Fourth Quarter\": {\"High\": 11.44, \"Low\": 9.47}, \"2019: -- Third Quarter\":
        {\"High\": 14.96, \"Low\": 10.26}, \"2019: -- Second Quarter\": {\"High\": 20.91, \"Low\": 12.61}, \"2019:
      -- First Quarter\": {\"High\": 18.19, \"Low\": 8.87}, \"2018: -- Fourth Quarter\": {\"High\": 12.16,
        \"Low\": 7.43}, \"2018: -- Third Quarter\": {\"High\": 20.6, \"Low\": 10.95}, \"2018: -- Second Quarter\":
      {\"High\": 18.3, \"Low\": 6.7}, \"2018: -- First Quarter\": {\"High\": 7.35, \"Low\": 6.0}}"
    },
    ...
]
\end{lstlisting}

%% file: appendix/prompt.tex
\section{Prompt Templates}
\label{app:prompt}

In this section, we present the core prompt templates used in our framework. To ensure reproducibility, we provide the full content of the system instructions and the specific action prompts used by the agents. Note that placeholders such as \texttt{\{dataset\_size\}} are dynamically filled during runtime.

\subsection{System Description}
The following system prompt is injected into the context of all agents (both One-Shot and Iterative) to define the AutoDataBench task environment, constraints, and evaluation criteria.

\begin{tcolorbox}[breakable, title=System Prompt: Task Instruction, colback=gray!5, colframe=gray!50!black, fonttitle=\bfseries]
You are participating data-driven model specialization, a new challenge designed to test your ability to act as an autonomous AI data synthesis scientist.

\textbf{===== TASK =====}\\
Your mission is to craft a high-quality, synthetic instruction-tuning dataset for a given task. 
Your formance is evaluated by how much a standard model's performance improves after being fine-tuned on the dataset you create. 
This task evaluates your core capabilities in understanding, synthesizing and distilling valuable training data from raw resources.
Please read the instructions below, which are specific to this environment. These instructions can be found again in `../agent/utils/instructions.py`.

\textbf{===== RESOURCES =====}\\
All necessary resources for your data synthesis process are located in \texttt{../data/}. This includes:
\begin{itemize}
    \item \texttt{../data/description.md}: Describes the target task, including the goal and example data format. [Description Content]
    \item \texttt{../data/seed/seed.json}: Necessary raw material from which you create instruction-tuning pairs, refine and expand this as needed. Provided only if required by the task. [Source Content]
    \item \texttt{../data/seed/few\_shot.json}: A small set of high-quality examples to follow for style, complexity, and format. [Few-shot Content]
    \item \texttt{../data/sample\_submission.json}: A template demonstrating the required JSON structure for your final submission instruction-tuning file. [Sample Submission]
    \item \texttt{../data/train\_config.yaml}: Contains the model architecture and default hyperparameters for fine-tuning. [Train Config]
\end{itemize}

\textbf{===== SUBMISSION =====}\\
Your final output is a synthetic training dataset. You MUST produce this dataset at EXACTLY \texttt{../submission/submission.json}, strictly adhere to the format specified in \texttt{sample\_submission.json}. The submission should be around \{dataset\_size\} entries, regardless of the amount of seed data provided (if any). Too little synthetic data severely degrades model training performance.

\textbf{===== ENVIRONMENT ======}\\
\begin{itemize}
    \item Teacher Models: API Access of \{teacher\_model\}.
    \item Target Models: Standard models to be fine-tuned and then evaluated: \{student\_model\}.
    \item GPU Avaliable: 160GB.
    \item Synthetic Data Constraint: We will filter your final submission to retain only entries matching the texttt{sample\_submission.csv} format with non-empty `output` fields, and use the first \{dataset\_size\} entries for training. 
    \item API Limit: You are limited to a total of \{api\_limit\} calls to the API-based teacher models.
    \item Runtime Limit: Your entire process must complete within \{max\_hours\} hours.
\end{itemize}

\textbf{===== IMPORTANT NOTES ======}\\
\begin{itemize}
    \item You must save your final dataset at exactly the specified path: \texttt{../submission/submission.json}. Save regularly to prevent data loss.
    \item Your only task is to generate instruction-tuning dataset. \textbf{Do not include any code for model training or evaluation.}
    \item You are only allowed to generate instruction-tuning data by calling the provided tearcher models. \textbf{Do not directly enumerate synthetic data in the code.}
    \item You should synthesize dataset that is diverse, complex, and task-aligned. The given few-shot examples are for reference only in terms of format and quality.
    \item You should be mindful to stay within your allocated API call quota. Avoid using loops that only check the number of generated entries while ignoring the API calls limit.
    \item You must ALWAYS prioritize calling the helper functions (if provided) directly to perform any relevant task. You are strictly prohibited from re-implementing their logic or creating any similar functions.
    \item You are participating in this competition independently. Ensure that the code you generate is DIRECTLY executable, no dummy implementations or placeholders of any kind.
\end{itemize}
\end{tcolorbox}

\subsection{One-Shot Agent}
The One-Shot Agent receives a single comprehensive prompt asking for a plan and the execution code. It does not receive feedback from the execution environment unless a retry is triggered by a crash.

\begin{tcolorbox}[breakable, title=One-Shot Agent: Generation Prompt, colback=gray!5, colframe=gray!50!black, fonttitle=\bfseries]
Your response should include a brief plan for the data synthesis, followed by a single markdown code block that implements this plan and generates the final synthetic data.
Conduct a concise analysis of the given information, and then wrap the plan and code separately in Markdown Code Blocks.

Example Response:\\
\texttt{[... Necessary Analysis ...]}\\
Here is the plan:\\
\texttt{[... Brief Plan ...]}\\
Here is the code:\\
\texttt{[... Implemented Code ...]}\\
\end{tcolorbox}

\subsection{Iterative Agent}
The Iterative Agent operates in a loop. Depending on the state of the previous iteration (Success, Execution Error, Submission Error, or Improvement Opportunity), it receives different prompts.

\begin{tcolorbox}[breakable, title=Iterative Agent: Draft (Initial Generation), colback=gray!5, colframe=gray!50!black, fonttitle=\bfseries]
\textbf{===== CURRENT STATUS =====}\\
- Remaining Time of All Iterations: \{remaining\_hours\} hours\\
- Remaining API Calls of All Iterations: \{remaining\_calls\}\\
- API Calls Limit For This Iteration: \{SESSION\_API\_LIMIT\}\\
Keep these constraints in mind when planning your next action.

Propose a brief plan and implementation code for synthesizing a high-quality dataset. Conduct a concise analysis of the given information, and then wrap the plan and code separately in \texttt{```} blocks.
\end{tcolorbox}

\begin{tcolorbox}[title=Iterative Agent: Debug (Execution Failure), colback=gray!5, colframe=gray!50!black, fonttitle=\bfseries]
The data generation process failed during execution. This is debug attempt \{debug\_attempts\}.\\
Failed Code:\\
\texttt{\{current\_code\}}\\
Error Message:\\
\texttt{\{last\_error\}}\\
Conduct a concise analysis of the given information, and then wrap the correction plan and the code in separate \texttt{```} blocks.
\end{tcolorbox}

\begin{tcolorbox}[breakable, title=Iterative Agent: Repair (Invalid Submission Format), colback=gray!5, colframe=gray!50!black, fonttitle=\bfseries]
Your previous attempt resulted in an invalid submission file located at \texttt{../submission/submission.json}.
Your task is to resolve the issue.

Here is your original plan: \texttt{\{current\_plan\}}\\
Here is your original code: \texttt{\{current\_code\}}\\
Here is the submission error details: \texttt{\{last\_error\}}

You can either repair the existing file or regenerate the data. Conduct a concise analysis... [Instructions on format]
\end{tcolorbox}

\begin{tcolorbox}[title=Iterative Agent: Improve (Optimization on Success), colback=gray!5, colframe=gray!50!black, fonttitle=\bfseries]
Your current best solution achieved a metric of \{best\_metric\}.
Your task is to improve the dataset. Make full use of the remaining API calls.
[Optional: Performance of the base model on this test set: \{score\}]

Here is your original plan: \texttt{\{best\_plan\}}\\
Here is your original code: \texttt{\{best\_code\}}\\
Here are the submission details: [Sample of submission file]

The model trained on that data failed on these cases:
\texttt{\{bad\_case\_sample\}}

Analyze these failures to identify model weaknesses and generate more targeted data. You can either improve the existing data quality or regenerate the data.
\end{tcolorbox}